\documentclass[journal]{IEEEtai}

\usepackage[colorlinks,urlcolor=blue,linkcolor=blue,citecolor=blue]{hyperref}

\usepackage{color,array}

\usepackage{amsmath}
\usepackage{mathtools}
\usepackage{mathrsfs}

\usepackage{subcaption}

\usepackage[noadjust]{cite}

\DeclareCaptionLabelSeparator{periodspace}{.\quad}
\usepackage{tabularx}
\usepackage{booktabs}

\mathchardef\mhyphen="2D 
\newcommand{\stimes}{{\times}}

\renewenvironment{IEEEbiography}[1]
  {\IEEEbiographynophoto{#1}}
  {\endIEEEbiographynophoto}

\begin{document}

\title{Bayesian Integration of Information Using Top-Down Modulated Winner-Take-All Networks}

\author{Otto van der Himst, Leila Bagheriye, and Johan Kwisthout}

\maketitle

\begin{abstract}
Winner-Take-All (WTA) circuits — a type of Spiking Neural Networks (SNN) — have been suggested as facilitating the brain’s ability to process information in a Bayesian manner. Research has shown that WTA circuits are capable of approximating hierarchical Bayesian models via Expectation-Maximization (EM). So far, research in this direction has focused on bottom-up processes. This is contrary to neuroscientific evidence that shows that, besides bottom-up processes, top-down processes too play a key role in information processing by the human brain. Several functions ascribed to top-down processes include direction of attention, adjusting for expectations, facilitation of encoding and recall of learned information, and imagery. This paper explores whether WTA circuits are suitable for further integrating information represented in separate WTA networks. Furthermore, it explores whether, and under what circumstances, top-down processes can improve WTA network performance with respect to inference and learning. The results show that WTA circuits are capable of integrating the probabilistic information represented by other WTA networks, and that top-down processes can improve a WTA network’s inference and learning performance. Notably, it is able to do this according to key neuromorphic principles, making it ideal for low-latency and energy-efficient implementation on neuromorphic hardware.
\end{abstract}

\begin{IEEEkeywords}
Neuromorphic Computing, Winner-take-all (WTA) circuit, hierarchical WTA network, Spiking Neural Network (SNN), spike-timing-dependent plasticity (STDP) learning, Bayesian inference, Top-Down Processes.
\end{IEEEkeywords}

\section{Introduction}\label{sec:introduction}

\IEEEPARstart{B}{ayesian} inference is one of the most prominent computational techniques in Artificial Intelligence (AI), being applied in a broad range of areas including statistical machine learning \cite{tipping_2004_bayesian, theodoridis_2015_machine}, causal discovery \cite{heckerman_2006_a-bayesian}, automatic speech recognition \cite{zweig_1998_speech}, spam filtering \cite{gomez_2006_content}, and clinical decision support systems \cite{sesen_2013_bayesian}. Bayesian inference uses new data in order to update existing models or hypotheses. The method relies on the well-known Bayes’ theorem:

\begin{equation}
    \overbrace{P(H|E)}^{Posterior} = \dfrac{\overbrace{P(E|H)}^{Likelihood} \overbrace{P(H)}^{Prior}}{\underbrace{P(E)}_{Evidence}}
\end{equation}

\noindent Where $H$ can be interpreted as the hypotheses or hidden causes, and $E$ as the available evidence or data. In essence this method can be used to continuously update the posterior probability over hypotheses based on the newly arriving evidence. While Bayesian inference in this manner yields an optimal estimation of the posterior, it quickly becomes intractable for all but the most simple environments. Addressing this issue, many methods have been conceived that approximate Bayesian inference. Such methods seek to obtain acceptable posterior estimations in a manner that is computationally tractable.

Besides interest from a purely AI perspective, there is considerable evidence that suggests that the human brain processes information in a Bayesian manner \cite{kersten_2004_object, kording_2006_bayesian, goldwater_2009_a-bayesian, shi_2016_development, haefner_2016_perceptual}. Even so, it remains an open question how the brain represents probabilistic information, how it performs Bayesian inference, and how it learns to represent Bayesian models correctly. 

\subsection{The Brain as Inspiration: Neuromorphic Computing}

While on the one hand, answering these questions is interesting because it tells us more about how the human brain works, it also serves as useful inspiration for field of neuromorphic computing. Recent years have seen an increasing growth in this field \cite{christensen_2022_2022}, where the computational principles of the traditional Von Neumann computer are substituted by more brain-inspired approaches. One of the most common neuromorphic approaches is to rely on Spiking Neural Network (SNNs). In its most simple form a SNN consists of (often sparsely) connected neurons. A neuron spikes when its membrane potential exceeds its threshold potential, causing a signal to be sent down its outgoing connections which excites or inhibits the connected neurons. This architecture can be adapted in many ways in order to change its behaviour and computational capabilities. The motivation for this can be to make the architecture more biologically plausible (e.g. by adopting a more complex neuron model), to make it more suitable for some practical purpose (e.g. by allowing neurons to send non-binary signals), or both.

Some of the most important features that are strived for in neuromorphic designs are computation based on locally available information and the co-location of computation and memory. Computation with local information entails that each computational component (e.g. a neuron in a SNN) requires access exclusively to local information (e.g. a neuron's 
own membrane potential, and spikes coming into its dendrites via connected axons). While this may seem like an undesirable restriction, such an architecture eliminates the need for constant communication with a global memory unit, which can serve as a major bottleneck in Von Neumann computers. Further, co-location of computation and memory refers to the parameters of the network being represented by the computational units themselves. That is, an inherent part of each computational unit (such as a neuron or an axon), is their own set of dynamic parameters (such as its membrane potential or a connection weight). These parameters reflect the state of their corresponding computational unit, and combined reflect the state of the network as a whole. Having memory as close as possible to where it is needed is a desirable property that reduces the time required for memory retrieval.

Relying on local rather than global information further opens the door for several additional desirable properties. For one, it facilitates event-based computation. In event-based computation, locality is leveraged such that each individual computational unit (e.g. a neuron) performs computations (e.g. spikes) exclusively when it receives some minimal amount of input. One might for example compare traditional cameras with the neuromorphic Dynamic Vision Sensor (DVS). When filming, a camera will shoot a given amount of frames each second, whereas a DVS (similar to the human visual system) will only register changes in its visual field. The event-based approach of the DVS has several advantages. First of all, between frames the traditional camera is essentially blind, meaning that it risks missing information when filming rapidly changing scenes. Contrary to this, the DVS responds to change regardless of its timing, and thus has no such temporal blind spots. Secondly, a camera will typically record a lot of redundant information. In a mostly static environment, each new frame captured will have a lot of information in common with the preceding frame. The DVS has no such issues given that it only responds to change, allowing for a very natural way of ignoring redundant data, and significantly reducing computational costs in many settings.

Local and event-based computation further facilitate a computer's capacity for parallel and asynchronous computation. This entails that each computational unit (e.g. a neuron) can act in parallel to many or all other computational units. Ideally, each unit acts purely based on local events that affect it (e.g. spikes travelling into a neuron's dendrites), and can completely ignore what happens outside of this, meaning that it can act or stay dormant without reference to some global computational unit. Under the right architecture, such a system is capable of processing many different stimuli at once, and of utilizing only those computational units that are necessary at a given time.

In general, the primary benefits of neuromorphic computers are considered to be a potential decrease in energy costs and response latency by orders of magnitude \cite{christensen_2022_2022}. This is reflected for example in the DVS when compared to a traditional camera: the DVS saves energy by responding only to changes, and reacts immediately to said changes rather than waiting for a frame to be shot. The hugely parallel and asynchronous potential of neuromorphic computers is also a key feature to be exploited.

In order to fully utilize these potential benefits two things are needed. First of all hardware is needed which functions according to these neuromorphic principles. Recent years have seen an increase in the design of such system, with notable chips including the Loihi \cite{davies_2018_loihi}, SpiNNaker, and BrainScaleS \cite{furber_2016_large} chips. Note that, for a variety of reasons, most chips do not exhibit all the neuromorphic properties mentioned above, instead choosing to focus on a subset of them. Secondly, the algorithms run on this hardware must adhere to neuromorphic principles. For example, an algorithm which requires constant access to global information will not be suitable for neuromorphic hardware. Likewise it is desirable for a SNN algorithm to be designed to spike very sparsely, given that this sparsity is what underlies the energy efficiency of neuromorphic designs.

\subsection{Neuromorphic solutions to Bayesian Inference} 

Circling back to Bayesian inference, we will proceed by highlighting several neuromorphic solutions to this problem. One direction concerns neural sampling methods. \cite{buesing_2011_neural} and \cite{pecevski_2011_probabilistic} propose a model for neural sampling that is on the one hand consistent with the dynamics of spiking neurons, and which on the other hand can also be understood from the perspective of probabilistic inference through Markov chain Monte Carlo (MCMC) sampling. Their method is similar to sampling approaches which have already been applied extensively (e.g. Boltzmann machines), moreover, the model is more biologically realistic as it incorporates aspects of inherent temporal dynamics and spike-based communication of a network of spiking neurons. \cite{huang_2016_bayesian} further provides a concrete neural implementation of this model that is capable both of approximate Bayesian inference and learning for any hidden Markov model.

Another direction relies on Winner-Take-All (WTA) circuits, a type of SNN that has been identified as a ubiquitous processing component in the brain \cite{nessler_2009_stdp, nessler_2013_bayesian, pecevski_2016_learning, guo_2019_hierarchical, yu_2020_emergent, yu_2020_sampling, bagheriye_2021_brain}. A WTA circuit is a simple SNN that consists of a single layer of excitatory neurons that is connected to a population of inhibitory neurons. Whenever an excitatory neurons fires, the population of inhibitory neurons is activated such that it sends back a strong inhibitory signal to all excitatory neurons. The excitatory neurons of the WTA circuit are themselves excited by a population of input neurons (e.g. sensory neurons). When combined with a Spike-Timing Dependent Plasticity (STDP) learning rule a WTA circuit can learn to distinguish between patterns of input activity. The intuition is that a WTA neuron firing in response to a particular input pattern will (due to STDP) become more likely to fire in response to this same input pattern in the future, while other neurons will not because they are inhibited and thus do not fire. Therefore if a WTA circuit is repeatedly exposed to structured input, a single neuron will become increasingly sensitive to a particular structure.

This process has been proven to be an approximation of the Expectation-Maximation (EM) algorithm \cite{nessler_2013_bayesian}. In this algorithm one can distinguish an expectation step and a maximization step. During the expectation step the network generates a posterior distribution over hidden causes of the current input given the current network weights. This is represented over time by the WTA excitatory neurons, where each excitatory spike represents a single stochastic sample from the posterior distribution encoded in the circuit. The maximization step maximizes the likelihood of the data in this distribution by updating the connection weights of the excitatory neurons according to a STDP learning rule.

Thus, a WTA circuit is capable first of all of generating a probabilistic representation of multinomial variables through spikes of its excitatory neurons. Secondly, it is able to infer the state of such a hidden variable from the input it receives from other neurons. And finally, it is able to learn the relation between patterns of input and hidden variables. Notably, it is able to do all this whilst adhering to all the neuromorphic principles that we mentioned earlier.

It has further been shown that multiple WTA circuits can be connected into a hierarchical WTA network to perform approximate mean field inference. \cite{guo_2019_hierarchical} introduced this approach to extend the method to be able to implement inference and learning for hierarchical Bayesian models. This method employs a fully factorized model to approximate a complex Bayesian model. This is particularly suitable for tree-structured models, though the neural implementation of such models can extend to arbitrary Bayesian networks by merging variables.

While WTA networks seem well suited for not just the processing of a single source of information, but for the integration of multiple sources of information, the latter has not been explicitly addressed in earlier research. We show that networks of WTA circuits are capable of executing this task. Specifically, we interpret  the spiking behaviour of a single WTA circuit (after learning according to STDP) as a compressed probabilistic representation of incoming information. As a WTA circuit can learn to distinguish between patterns of raw sensory neuron spikes, we show that a WTA circuit can learn to distinguish between patterns in the spiking behaviour of preceding WTA circuits, such that it integrates their information into a single probabilistic representation.

As we add more layers to the WTA network, it becomes important to take into account not just bottom-up processes, but also top-down processes. Bottom-up processes concern flows of information starting with spikes from sensory neurons, such as from neurons in the retina. In the human brain this bottom-up activation flows from the sensory neurons up through a hierarchy of processing layers (e.g. visual areas V1-V5); in addition to these bottom-up processes, however, there are top-down processes in which activation flows in the other direction (e.g. from V5 back to V1) \cite{gilbert_2013_top-down, harris_2013_cortical, dijkstra_2017_distinct, semedo_2021_feedforward}. There are several functions ascribed to these top-down processes, including directing of attention, adjusting for expectations, adjusting for the perceptual task, facilitation of encoding and recall of learned information, and imagery. Specifically for visual tasks these processes are thought to play an important role in perceptual grouping, perceptual constancies, contour integration, surface segmentation, and shape recognition \cite{gilbert_2013_top-down}.

In \cite{nessler_2013_bayesian} only a single WTA circuit is considered, rather than a network of circuits. Brain research shows that top-down processes do not extend to sensory neurons, and since the design of \cite{nessler_2013_bayesian} includes no other neuron layers, it cannot include top-down processes. In the work of \cite{guo_2019_hierarchical} where an additional layer of WTA circuits is added to the network, top-down processes do become possible. Indeed \cite{guo_2019_hierarchical} mentions that the connections between the two WTA layers are bidirectional, thereby allowing feedback, however, further explanation on the role or impact of this feature is still missing. In our work we extend the hierarchical network of \cite{guo_2019_hierarchical} by adding a WTA circuit that integrates information from multiple hierarchical networks. The addition of this layer further increases the potential impact of top-down processes.

The purpose of our work then is twofold. First we explore experimentally whether WTA circuits can chain together separate WTA networks into a larger WTA network, with beneficial consequences to its inference and learning capacities. We demonstrate that WTA circuits are in fact suitable for such a task. Secondly, we explore the role that top-down processes have in WTA networks. On the one hand this is done by demonstrating that top-down processes are able to improve a WTA network's capacity to represent variables, perform inference, and learn. On the other hand it is done by demonstrating that this effect is greater in larger WTA networks (i.e. in our integration design as compared to the hierarchical design of \cite{guo_2019_hierarchical}). Together, this research highlights the feasibility of WTA networks as a fully neuromorphic (i.e. local, event-based, parallel, asynchronous) approach to performing Bayesian inference. The rest of the paper is organized as follows. Section \ref{sec:model} provides a formal definition of a WTA circuit and the underlying Bayesian model it represents. Section \ref{sec:experiments} describes our experimental setup and reports the experimental results. And finally, the conclusions are drawn in Section \ref{sec:conclusion}.

\section{WTA network definition and underlying Bayesian model}\label{sec:model}

We will now define in formal terms what a WTA circuit is, how it can represent a Bayesian model, and how it can be extended to a WTA network consisting of multiple circuits. Table \ref{tab:notation} provides an overview of the mathematical notation used throughout this paper.

\begin{table}[h]
\setlength{\tabcolsep}{3pt}

\caption{WTA and Bayesian network notation}
\begin{tabular*}{21pc}{p{56pt}p{180pt}}
\toprule
  
\multicolumn{2}{l}{{\underline{WTA circuit}:}}                                       \\

$\pmb{x}$ & observable variables $x_1, ..., x_V$ \\

$\pmb{s}$ & sensory neurons $s_1, ..., s_W$ encoding observable variables $\pmb{x}$ \\

$\pmb{y}$ & neurons $\{\pmb{y}^\uparrow, \pmb{y}^\downarrow\}$ providing excitatory input to neurons $\pmb{z}$ \\
$\pmb{y}^\uparrow$ & neurons $y^\uparrow_1, ..., y^\uparrow_M$ providing bottom-up excitatory input to neurons $\pmb{z}$ \\
$\pmb{y}^\downarrow$ & neurons $y^\downarrow_1, ..., y^\downarrow_N$ providing top-down excitatory input to neurons $\pmb{z}$ \\

$\pmb{z}$ & WTA circuit excitatory neurons $z_1, ..., z_K$ \\

$\pmb{w}$ & all synaptic weights $\{\pmb{w}^\uparrow, \pmb{w}^\downarrow\}$ \\
$\pmb{w}^\uparrow$ & bottom-up synaptic weights $\left\{w^\uparrow_{km} | k \in \{1, ..., K\}, m \in \{1, ..., M\}\right\}$ \\
$\pmb{w}^\downarrow$ & top-down synaptic weights $\left\{w^\downarrow_{kn} | k \in \{1, ..., K\}, n \in \{1, ..., N\}\right\}$ \\
$\Delta w(t)$ & the change of a synaptic weight at time t \\

$t$ & a particular point in time \\
$t^f$ & a particular point in time at which an excitatory neuron from $\pmb{z}$ fired \\

$\mu_k(t)$ & membrane potential of neuron $z_k$ at time $t$ \\
$u_k(t)$ & excitatory input to neuron $z_k$ at time $t$ \\
$I(t)$ & combined (scalar) inhibitory signal $I^l(t) + I^c(t)$ sent to all neurons $\pmb{z}$ at time $t$ \\
$I^l(t)$ & lateral inhibition signal used to drive competition between neurons $\pmb{z}$ \\
$I^c(t)$ & inhibitory signal used to control and stabilize over time the excitatory signal $\pmb{u}$ sent to neurons $\pmb{z}$ \\

$r_k(t)$ & firing rate of neuron $z_k$ (which fires according to an inhomogeneous Poisson process) \\
$R(t)$ & combined firing rate of all neurons in $\pmb{z}$ (which collectively fire according to an inhomogeneous Poisson process) \\

$z_k(t)$ & 1 if neuron $z_k$ fired at time $t$, 0 otherwise; likewise for neurons $\pmb{s}$ and $\pmb{y}$ \\

\multicolumn{2}{l}{{\underline{WTA network}:}} \\
$G$ & number of WTA circuits in the WTA network \\
$\pmb{z}^g$ & circuit $g \in\{1,...,G\}$ of the WTA network; likewise for other circuit components \\
$\pmb{X}$ & collection of all network observable variables ${\pmb{x}^1, ..., \pmb{x}^G}$ \\
$\pmb{S}$ & collection of all network sensory neurons ${\pmb{s}^1, ..., \pmb{s}^G}$ \\
$\pmb{Y}$ & collection of all network inputs ${\pmb{y}^1, ..., \pmb{y}^G}$ \\
$\pmb{Z}$ & collection of all network excitatory neurons ${\pmb{z}^1, ..., \pmb{z}^G}$ \\
$\pmb{W}$ & collection of all network weights ${\pmb{w}^1, ..., \pmb{w}^G}$ \\

\multicolumn{2}{l}{{\underline{Bayesian Model}:}}                                       \\ 
$\theta$ & generative model parameters \\
$k$ & hidden variable considered to be an underlying cause of observable variables \pmb{x}  \\
$\pmb{k}$ & hidden variables $k^1, ..., k^G$ considered to be the underlying causes of observable variables \pmb{x}  \\
$\hat{\pmb{y}}$ & models the distribution of $\pmb{y}(t)$ over all points in time \\
$z_k$ & is 1 if the hidden cause is $k$, 0 otherwise \\

\bottomrule
\end{tabular*}
\label{tab:notation}
\end{table}

\subsection{WTA Circuit Definition}

In our discrete-time model, a WTA circuit consists of a layer $\pmb{z} = \{z_1, ..., z_K\}$  of $K$ excitatory integrate-and-fire neurons. Like \cite{nessler_2013_bayesian} we adopt a stochastic firing model in which the firing probability of each neuron $z_k$ depends exponentially on the membrane potential $\mu_k$ of said neuron. The membrane potential $\mu_k$ of each neuron $z_k$ is updated at each time step as a function of its current state and of incoming excitatory and inhibitory signals.

\subsubsection{Excitation}
Excitatory inputs consist of bottom-up inputs generated by a population of $M$ neurons $\pmb{y}^\uparrow = \{y^\uparrow_1, ..., y^\uparrow_M\}$, and – if top-down processes are enabled – additionally include top-down inputs generated by a population of $N$ neurons $\pmb{y}^\downarrow = \{y^\downarrow_1, ..., y^\downarrow_N\}$. In our work, neurons $\pmb{y}^\uparrow$ are either sensory neurons $\pmb{s} = \{s_1, ..., s_W\}$ or neurons $\pmb{z}^\prime$ from preceding WTA circuits, while neurons  $\pmb{y}^\downarrow$ are exclusively neurons $\pmb{z}^\prime$ from successive WTA circuits. 

Input neurons $y^\uparrow$ and $y^\downarrow$ have outgoing excitatory connections leading to neurons $\pmb{z}$, the strength of the connections is expressed by weights $\pmb{w}^\uparrow = \{w^\uparrow_{km} | k \in \{1, ..., K\}, m \in \{1, ..., M\}\}$ and $\pmb{w}^\downarrow = \{w^\downarrow_{kn} | k \in \{1, ..., K\}, n \in \{1, ..., N\}\}$. If we denote $y(t)$ to mean that input neuron $y$ fired at time $t$, then we can define the combined strength of the excitatory inputs for neuron $z_k$ at time $t$ to be:
\begin{equation} \label{eq:excitation}
    u_k(t) = \overbrace{\sum_{m=1}^M w^\uparrow_{km} y^\uparrow_m(t)}^{Bottom-up\ excitation} + \overbrace{\sum_{n=1}^N w^\downarrow_{kn} y^\downarrow_n(t)}^{Top-down\ excitation}
\end{equation}

\subsubsection{Inhibition}
At every point in time, each neuron $z_k$ is further influenced by an identical (scalar) inhibitory signal $I(t) = I^l(t) + I^c(t)$. The role of this inhibitory signal is twofold. First of all, inhibition signal $I^l(t)$ installs a mechanism of lateral inhibition which drives competition between neurons $\pmb{z}$. Secondly, inhibition signal $I^c(t)$ is used to exert control over the combined input signal $u_k(t) - I(t)$, which has several purposes that will be elaborated on in section \ref{sec:discrete-time-implementation}; for now we will assume that $I^c(t) = 0$.

Lateral inhibition is realized by connecting all neurons $\pmb{z}$ to a population of inhibitory neurons. These neurons send an inhibitory signal to all neurons $\pmb{z}$ according to:
\begin{equation}
    I^l(t) =
    \begin{cases}
    \infty & \text{if } \pmb{z}(t-1)\\
    0,              & \text{otherwise}
    \end{cases}
\end{equation}
Meaning that at any time point $t$ that a neuron in $\pmb{z}$ fires, this population of neurons is activated, causing it to send back a strong inhibitory signal that –-- in our model –-- is strong enough to reset the membrane potential of all neurons $\pmb{z}$ to zero. This mechanism of lateral inhibition, combined with STDP, allows a neuron $z_k$ to become distinctly sensitive to recurring input spiking patterns, a process which we will elaborate on in section \ref{sec:EM}.

\subsubsection{Membrane potential}
The membrane potential $\mu_k$ of a neuron $z_k$ is thus updated at each timestep as a function of its current state, and of incoming excitatory and inhibitory signals, as according to:
\begin{equation} \label{eq:membrane-potential}
    \mu_k(t) = \max{(0, \mu_k(t-1) + u_k(t) - I(t))}
\end{equation}
Where $\mu_k(0) = 0$. We can then define the probability of a neuron $z_k$ spiking at time $t$ to depend exponentially on its membrane potential:

\begin{figure}
\centerline{\includegraphics[width=18.5pc]{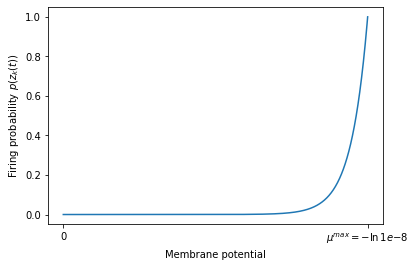}}
    \caption{Firing probability $p \bigl( z_k(t) \bigr) = \exp \bigl( \mu_k(t) - \mu^{max} \bigr)$ of each neuron $z_k$ as a function of its membrane potential, with $\mu^{max} = {-}\ln1e{-}8$.}
    \label{fig:firing-probability}
\end{figure}

\begin{equation} \label{eq:spike-proportional}
    p \bigl( z_k(t) \bigr) \propto e^{\mu_k(t)}
\end{equation}

\subsection{Underlying Bayesian Model} \label{sec:bayesian-model}

Having defined the components that make up a WTA circuit, we will now illustrate how it can be interpreted as a Bayesian model. As described in \cite{nessler_2013_bayesian}, each neuron $z_k$ fires according to an independent inhomogeneous Poisson process whose firing rate is given by:
\begin{equation}
    r_k(t) = e^{\mu_k(t)} 
\end{equation}
The combined firing rate of neurons $\pmb{z}$ can then be expressed as the sum of K independent Poisson processes, resulting in an inhomogeneous Poisson process with a rate $R(t)$ such that:
\begin{equation}
    R(t) = \sum_{k=1}^K r_k(t) 
\end{equation}
Moreover, in continuous time, during any infinitesimally small time interval $[t, t+\delta t]$, each neuron $z_k$ spikes with probability $r_k(t)\delta t$. Therefore, if at a given time $t$ (i.e. within $[t, t+\delta t]$) a single neuron in $\pmb{z}$ spikes, the conditional probability $q_k(t)$ that this spike was generated by neuron $z_k$ is:
\begin{equation} \label{eq:instantaneous-spike-distribution}
    q_k(t) = \dfrac{r_k(t) \delta t}{R(t) \delta t} = \dfrac{e^{\mu_k(t)}}{\sum_{k^\prime=1}^K e^{{\mu_{k^\prime}}(t)}}
\end{equation}
This holds in continuous time, where the probability of two neurons spiking at the exact same time is zero. In discrete time, where multiple neurons can spike simultaneously, additional conditions need to be satisfied in order to approximately arrive at equation \ref{eq:instantaneous-spike-distribution}, we elaborate on this in section \ref{sec:discrete-time-implementation}.

We can interpret the spike distribution defined by $q_k(t)$ as a generative model over multinomial observed variables $\pmb{x} = \{x_1, ..., x_V\}$ and hidden cause $k$, parametrized by $\theta$:
\begin{equation}\label{eq:generative-model}
	p(k, \pmb{x} | \theta) = p(k | \theta) \prod_{v=1}^V p(x_v|k, \theta)
\end{equation}
For one, such a model can be used to generate observable variables $x_1, ..., x_V$ by sampling from $k$ from prior distribution $p(k | \theta)$. For another, by applying Bayes’ rule, it can be used to approximate the posterior distribution:
\begin{equation}\label{eq:posterior}
    p(k | \pmb{x}, \theta) \propto p(k | \theta) p(\pmb{x} | k, \theta)
\end{equation}
and thus infer the hidden cause $k$ of the observation $\pmb{x}$.

In order to link equations \ref{eq:instantaneous-spike-distribution} and \ref{eq:posterior}, we define population codes to represent observable variables $\pmb{x}$ by a set of input neurons $\pmb{y} = \{\pmb{y^\uparrow}, \pmb{y^\downarrow}\}$, and hidden variable $k$ by circuit neurons $\pmb{z}$. Observable variables $\pmb{x}$ are encoded such that for each possible value of each variable $x_v$, there is exactly one neuron in $\pmb{y}$ that encodes it. Likewise, each of the K possible values that hidden variable $k$ can assume is represented by exactly one neuron in $\pmb{z}$.

We define $\pmb{y}(t)$ to be the activation of neurons $\pmb{y}$ at time $t$ (i.e. which input neurons fired at time t). Further, $\hat{\pmb{y}}$ represents a variable in the probabilistic model that models the distribution of $\pmb{y}(t)$ over all points in time. 
We can the neuron population codes defining binary variable vectors $\pmb{y}$ and $\pmb{z}$ to reformulate the probabilistic model $p(k, \pmb{x} | \theta)$ as:
\begin{equation}
	p(\pmb{z}, \hat{\pmb{y}} | \pmb{w}) = 
	\dfrac{1}{Z} \sum_{k=1}^K z_k \exp\left(\sum_{m=1}^M w_{km} \hat{y}_m \right)
\end{equation}
Where $z_k = 1$ if the hidden cause is $k$ and $z_k = 0$ otherwise, and where Z is the normalization constant.

This generative probabilistic model can then be described in terms of a WTA circuit SNN model. Evaluating the network at each time point $t^f$ at which a neuron in $\pmb{z}$ fires, we can compute the posterior probability of cause k by applying Bayes' rule to $p(\pmb{z}, \hat{\pmb{y}} | \pmb{w})$, arriving at\footnote{\cite{nessler_2013_bayesian} uses weights $w_{k0}$, representing the excitability of a neuron $z_k$, to model the prior probability of hidden cause $k$. For brevity we instead assume at all times a uniform prior (which fits with the dataset used in our experiments).}:
\begin{equation} \label{eq:spike-posterior}
    \begin{aligned}
        p(k | \pmb{y}(\pmb{t}^{\prime}, t^f), \pmb{w}) &= 
        \dfrac{\overbrace{e^{\mu_k(t^f-1) + \sum_{m=1}^M w_{km} y_m(t^f) - I(t^f)}}^{\text{likelihood } {p(\pmb{y}(\pmb{t}^{\prime}, t^f) | k, \pmb{w})}}} 
        {\underbrace{\sum_{k^{\prime}=1}^K e^{\mu_{k^{\prime}}(t^f-1) + \sum_{m=1}^M w_{{k^{\prime}}m} y_m(t^f) - I(t^f)} }_{p(\pmb{y}(\pmb{t}^{\prime}, t^f) | \pmb{w})}} \\
        &= \dfrac{e^{\mu_k(t^f)}}{\sum_{k^\prime=1}^K e^{{\mu_{k^\prime}}(t^f)}} = q_k(t^f)
    \end{aligned}
\end{equation}
where $\pmb{y}(\pmb{t}^{\prime}, t^f)$ concerns the spiking history of neurons $\pmb{y}$ from $t^f$ back to the time point following the most recent non-zero lateral inhibition signal. Thus at all time points $t^f$ a spike from a neuron $z_k$ can be seen as a sample from the posterior distribution $p(k | \pmb{y}(\pmb{t}^{\prime}, t^f), \pmb{w})$. 

\subsection{Discrete time implementation} \label{sec:discrete-time-implementation}

In discrete time, spiking probabilities do not naturally follow equation \ref{eq:instantaneous-spike-distribution} like they do in continuous time. In our model, each neuron $z_k$ fires at time $t$ with probability:
\begin{equation}
    p(z_k(t)) = e^{\mu_k(t) - \mu^{max}}
\end{equation}
where $\mu^{max}$ is the maximum membrane potential of neurons $\pmb{z}$, at which the probability of firing is one (see Fig. \ref{fig:firing-probability}). Note that this means that the firing probability of each neuron $z_k$ does not depend directly on the membrane potentials of the other neurons in $\pmb{z}$. Further, it does not include direct information of the firing rates of other neurons in $\pmb{z}$, nor of the desired combined firing rate of neurons $\pmb{z}$. In order to still arrive approximately at the distribution described by \ref{eq:instantaneous-spike-distribution}, we pass on this information indirectly through inhibition signal $I^c(t)$.

A second role of inhibition signal $I^c(t)$ is related to variations in the excitatory input signal. In discrete time it is possible for multiple neurons in $\pmb{y}$ to fire simultaneously. Furthermore, the number of neurons in $\pmb{y}$ that fire simultaneously can vary greatly between time steps. Given the exponential relation between a neuron's membrane potential and its spiking probability, sudden bursts in excitatory input can cause the spiking probability of multiple neurons in $\pmb{z}$ to go from close to zero, to one, from one timestep to the next. In this scenario, no distinction is made between the excitatory signals being received by each of these neurons. In order to avoid this, inhibition signal $I^c(t)$ is used to balance variation in the excitatory signals received over time.

The role of inhibition signal $I^c(t)$ is thus to pass on to neurons $\pmb{z}$, approximate information about the present firing rates of neurons $\pmb{z}$, as well as information about the desired combined firing rate of said neurons. In this manner, the signal can be used to exert control over the combined firing rate of neurons $\pmb{z}$, such that it remains stable over time and is independent of the fluctuation of activity in $\pmb{y}$. At each time step, the signal must be strong enough such that it prevents the spiking probabilities of neurons $\pmb{z}$ from exploding. Further, it must not be so strong that it completely overrides excitatory inputs, in which case information sent by neurons $\pmb{y}$ would be lost.

This mechanism is included in our model through a second population of inhibitory neurons which at every timestep sends an inhibitory signal to all neurons $\pmb{z}$. The strength of this signal is defined to be:
\begin{equation}
    I^c(t) = \psi|\pmb{y}(t)|
\end{equation}
Where $|\pmb{y}(t)|$ is the number of input neurons that fired at time $t$. And where $\psi$ is a scalar variable that changes every timestep as a function of the divergence of overall firing of $\pmb{z}$ from the desired firing rate.

\subsection{WTA Network Definition}

\begin{figure*}%
\centering
\begin{subfigure}{.45\textwidth}
\includegraphics[width=\columnwidth]{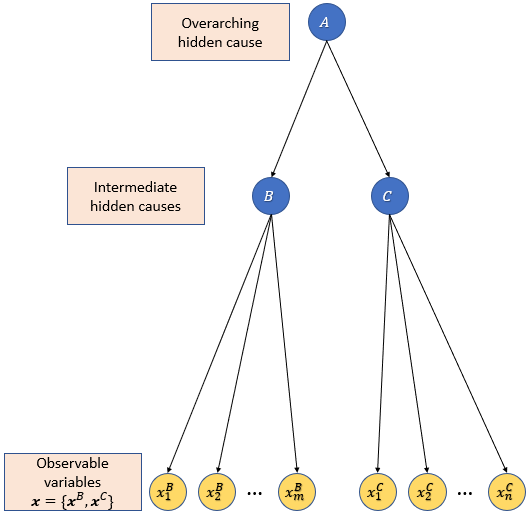}%
\caption{}%
\label{fig:abstract-model}%
\end{subfigure}\hfill%
\begin{subfigure}{.55\textwidth}
\includegraphics[width=\columnwidth]{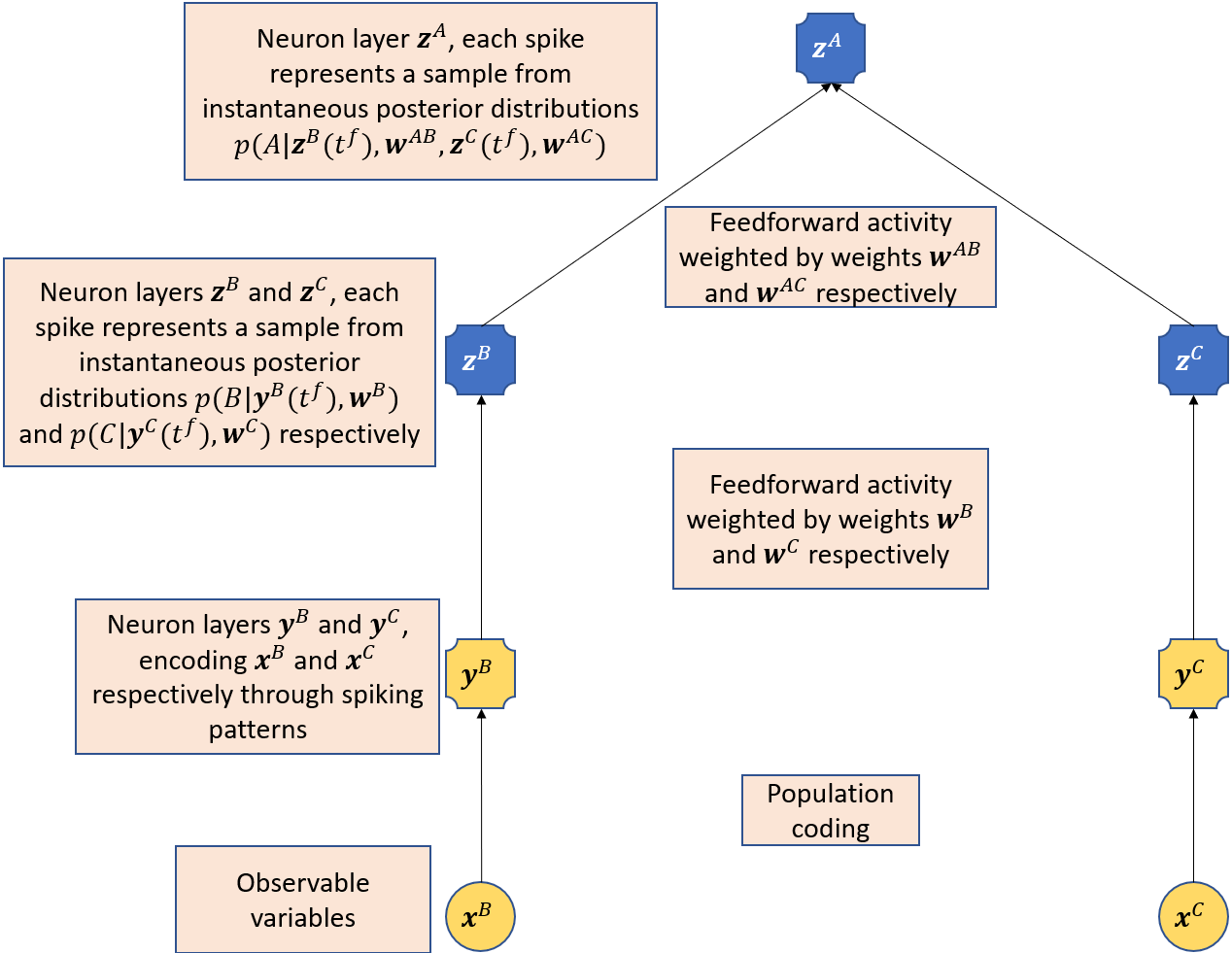}%
\caption{}%
\label{fig:abstract-wta}%
\end{subfigure}\hfill%

\begin{subfigure}{.5\textwidth}
\includegraphics[width=\columnwidth]{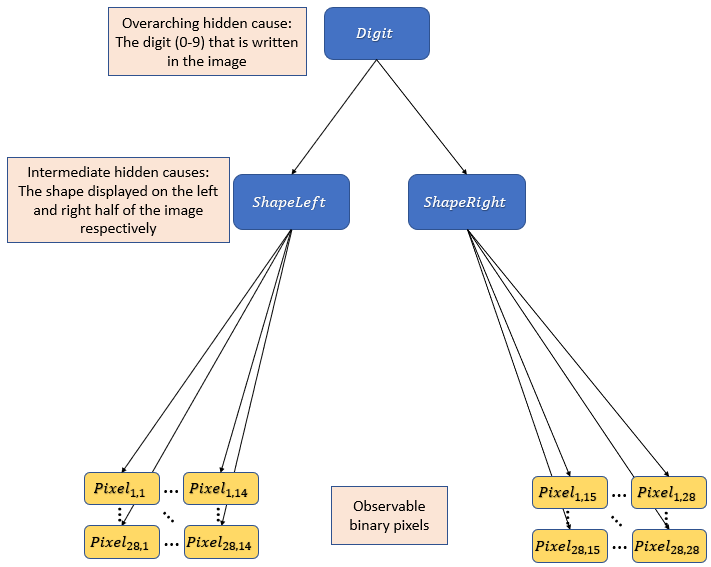}%
\caption{}%
\label{fig:concrete-model}%
\end{subfigure}\hfill%
\begin{subfigure}{.5\textwidth}
\includegraphics[width=\columnwidth]{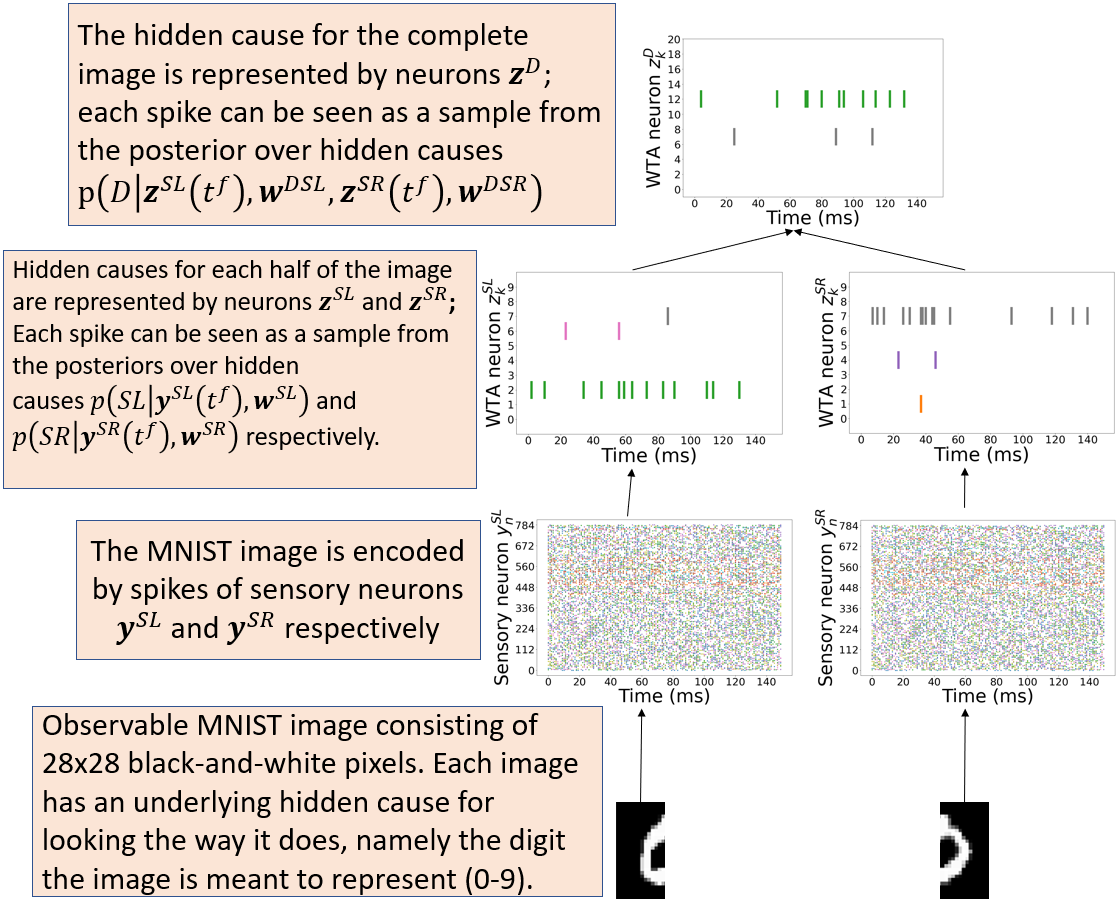}%
\caption{}%
\label{fig:concrete-wta}%
\end{subfigure}%
\caption{
Abstract and concrete examples of Bayesian models and corresponding WTA network. (a) Simple abstract Bayesian model. (b) WTA network that encodes abstract hidden variables $A$, $B$, and $C$. (c) Bayesian model of underlying causes of MNIST images. (d) WTA network that encodes underlying causes of MNIST images. Note that while the example models and networks are simple, the principles highlighted here hold for any tree-structured model, regardless of the amount of layers and the amount of variables per layer. Top-down processes are included by adding reversed connections with their own separate weights between each pair of connected WTA circuits.}
\label{fig:model-wta-examples}
\end{figure*}

Circuit neurons $\pmb{z}$ can be used as input for other circuits without changing the dynamics. Thus if we have a set of $G$ WTA circuits with neurons $\pmb{Z} = \{\pmb{z}^1, ..., \pmb{z}^G\}$, we can organize these in a hierarchical network such that each WTA circuit receives feedforward input from neurons $\pmb{z}$ from an arbitrary number of WTA circuits, and has its own neurons $\pmb{z}$ send feedforward activity to at most one WTA circuit. In addition we include feedback connections with their own separate weights to model top-down processes, thus allowing WTA circuits to furthermore receive input from other circuits that are one step higher in the hierarchy. Note that within a WTA network, a layer of neurons can have the role of excitatory neurons $\pmb{z}$ with respect to one WTA circuit, while playing the role of input neurons $\pmb{y}$ with respect to another WTA circuit. Since the probability distribution of a WTA circuit is solely dependent on input from neurons $\pmb{y}$, we can define hierarchical WTA by the joint probability distribution:
\begin{equation}
    p(\pmb{k} | \pmb{Y}(t^f), \pmb{W}) = \prod_{g=1}^{G} p(k^g | \pmb{y}^g(t^f), \pmb{w}^g)
    \label{eq:joint-wta}
\end{equation}
Where $\pmb{k} = \{k^1, ..., k^G\}$ are the underlying causes encoded respectively by neuron layers $\pmb{z}^1, ..., \pmb{z}^G$, where $\pmb{Y} = \{\pmb{y}^1, ..., \pmb{y}^G\}$ is the combined inputs to all circuits, and where $\pmb{W} = \{\pmb{w}^1, ..., \pmb{w}^G\}$ is the combined weights of all circuits. Fig. \ref{fig:model-wta-examples} illustrates both in an abstract and concrete manner how, following this formulation, a Bayesian model can be represented by a network of WTA circuits.

\subsection{Expectation Maximization through STDP} \label{sec:EM}

We have shown how WTA networks can represent generative probabilistic models. In addition to this, WTA networks are capable of learning the parameters of such a model. As was mentioned earlier, the combination of lateral inhibition and a Spike-Timing-Dependent Plasticity (STDP) learning rule, allows a neuron $z_k$ to become distinctly sensitive to common input spiking patterns. As is described in \cite{nessler_2013_bayesian} a WTA circuit approximates a stochastic, online version of the Expectation Maximization (EM) algorithm when it adopts a particular STDP rule. \cite{nessler_2013_bayesian} names this principle SEM (spike-based EM). The rule adopted by \cite{nessler_2013_bayesian}, as well as by \cite{guo_2019_hierarchical} and ourselves, is the biological STDP rule:

\begin{equation}
    \Delta w_{kn}(t) = \alpha(t^\mathit{diff}) c e^{-w_{kn}} - 1
\end{equation}
 
\noindent Where $c$ is a constant, where $t^\mathit{diff}$ marks the time difference between the most recent post-synaptic spike of neuron $z_k$ and pre-synaptic spike of neuron $y_n$, and where $\alpha(t^\mathit{diff})$ is shaped according to an alpha shaped kernel:

\begin{equation}
  \begin{aligned}[b]
    & \alpha(t^\mathit{diff}) \\
    & = \dfrac{1}{\tau_f - \tau_s} \left( \exp \left(-\dfrac{t^\mathit{diff}}{\tau_f}\right) - \exp \left(-\dfrac{t^\mathit{diff}}{\tau_s}\right) \right) \Theta(t^\mathit{diff})
\end{aligned}
\end{equation}

\noindent An update $\Delta w_{kn}(t)$ is finally weighed by adaptive learning rate $\eta_k(t)$ that over time diminishes the weight updates for connections coming into neuron $z_k$ proportionally to how often $z_k$ has spiked.
 
\begin{figure}
\centerline{\includegraphics[width=18.5pc]{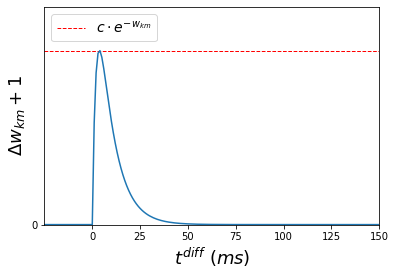}}
    \caption{STDP learning curve. The graph shows how the strength of weight change $\Delta w_{km}$ varies as a function of the difference $t^\mathit{diff}$ between the most recent post- and pre-synaptic spike times.}
    \label{fig:stdp}
\end{figure}

In essence, this rule causes weights of connections going from pre-synaptic neurons $\pmb{y}$ to a neuron $z_k$ to be updated every time $z_k$ fires. If a pre-synaptic neuron fired within some brief timespan (regulated by constants $\tau_f$ and $\tau_s$) before $z_k$ fired, then the connection weight between the two is increased, if not, then it is decreased. The strength and direction of the change is dependent on the relative timing of the pre- and post-synaptic spike, on the strength of the connection weight before the update, and on constants $\tau_f$ and $\tau_s$. Heavyside step function $\Theta(t^\mathit{diff})$ ensures that the function returns zero for negative values of $t^\mathit{diff}$. Fig. \ref{fig:stdp} visualizes STDP weight updates for $\tau_f=2$ and $\tau_s=8$.

This dynamic has the effect that when a neuron $z_k$ fires, it becomes more sensitive to pre-synaptic neurons that fired in the right time window before $z_k$ did, whereas it becomes less sensitive to pre-synaptic neurons that fired outside this time window. Thus, in the future this neuron will be more likely to respond to a particular pattern of pre-synaptic spikes, and if it does again respond to the same pattern, this is reinforced once more. At the same time, lateral inhibition ensures that when a neuron $z_k$ fires in response to some pattern of pre-synaptic spikes, others are inhibited and do not fire. As such they are forced to respond to a different pattern of pre-synaptic spikes, thus ensuring that each neuron becomes sensitive to its own distinct pattern (Fig. \ref{fig:weights-visualization} visualizes this process).

As shown by \cite{nessler_2013_bayesian} this can be interpreted within the framework of EM, where one can identify the expectation step and the maximization step in the circuit dynamics. The expectation step is represented by spikes from neurons $\pmb{z}$, where each spike can be considered to be a sample from the currently encoded posterior distribution over hidden causes. The maximization step is represented by the STDP weight update following a spike. This change of connection weights optimizes the posterior distribution they encode according to evidence provided by the spike. In this manner a WTA circuit is interpreted as a stochastic online EM algorithm.

\section{Experiments}\label{sec:experiments}

\begin{figure*}[hbpt!]
    \centering
    \includegraphics[width=\textwidth]{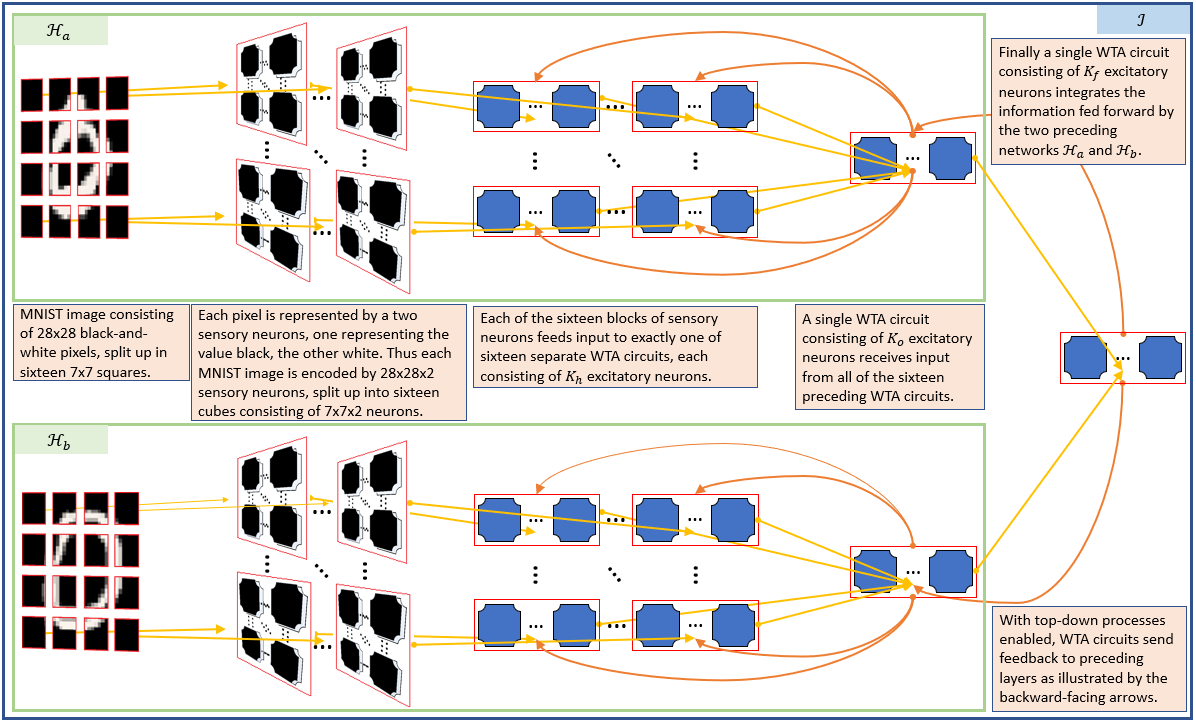}
    \caption{Visualization and description of our proposed integration network design ($\mathcal{I}$) that processes two separate input sources through the integration of information fed forward by two hierarchical networks ($\mathcal{H}_a$ and $\mathcal{H}_b$).}
    \label{fig:network-design}
\end{figure*}

We performed two experiments. The first experiment concerns a comparison between the two-layer hierarchical network design of \cite{guo_2019_hierarchical}, and our own three-layer integration design. In the second experiment we include top-down processes, and observe how these impact network performance.

\subsection{Experimental Setup}

Like \cite{nessler_2013_bayesian} and \cite{guo_2019_hierarchical} we assess the performance of network designs with respect to the MNIST dataset. In essence this becomes an unsupervised learning task, where the network learns solely through exposure to MNIST images, and the already described STDP dynamics, to distinguish between different input patterns, in this case black-and-white\footnote{The MNIST dataset in fact contains grayscale images, which we convert to black-and-white by converting all values that are not completely white to the value black.} images of handwritten digits 0-9.

The MNIST images are 28x28 pixel black-and-white images, which we encode in the same manner as \cite{guo_2019_hierarchical}. Each pixel is encoded by two neurons, one neuron represents the pixel value black, the other white. Thus the input layer of (sensory) neurons consists of 28x28x2 neurons. An active neuron fires according to a Poisson process, an inactive neurons remains silent. Thus each MNIST image is encoded over time by 784 Poisson-spike trains which we consider to be sensory input. In our experiments we present each MNIST digit for 150 ms, and set each active sensory neuron to adopt a firing rate of 200 Hz.

In the design of \cite{guo_2019_hierarchical} --- which we will refer to as the hierarchical design --- the first layer consists of 16 WTA circuits with each $K_h$ excitatory neurons. Each of these circuits receive input from a separate 7x7x2 cube of sensory neurons. The activation of each of these WTA circuits is then fed forward to a single WTA circuit with $K_o$ excitatory neurons that makes up the final layer.

Our own design is an extension of the hierarchical design. In our design --- which we will refer to as the integration design --- the final layers of two identical hierarchical networks ($\mathcal{H}_a$ and $\mathcal{H}_b$) are connected to one another by an additional WTA circuit with $K_f$ excitatory neurons, which comprises the final layer of the design. This integration network ($\mathcal{I}$) is thus capable of separately processing two sets of stimuli (one by $\mathcal{H}_a$ and the other by $\mathcal{H}_b$) before integrating the information into a single representation. The design is further illustrated in Fig. \ref{fig:network-design}.

\begin{figure*}%
\centering
\begin{subfigure}{.32\textwidth}
\includegraphics[width=\columnwidth]{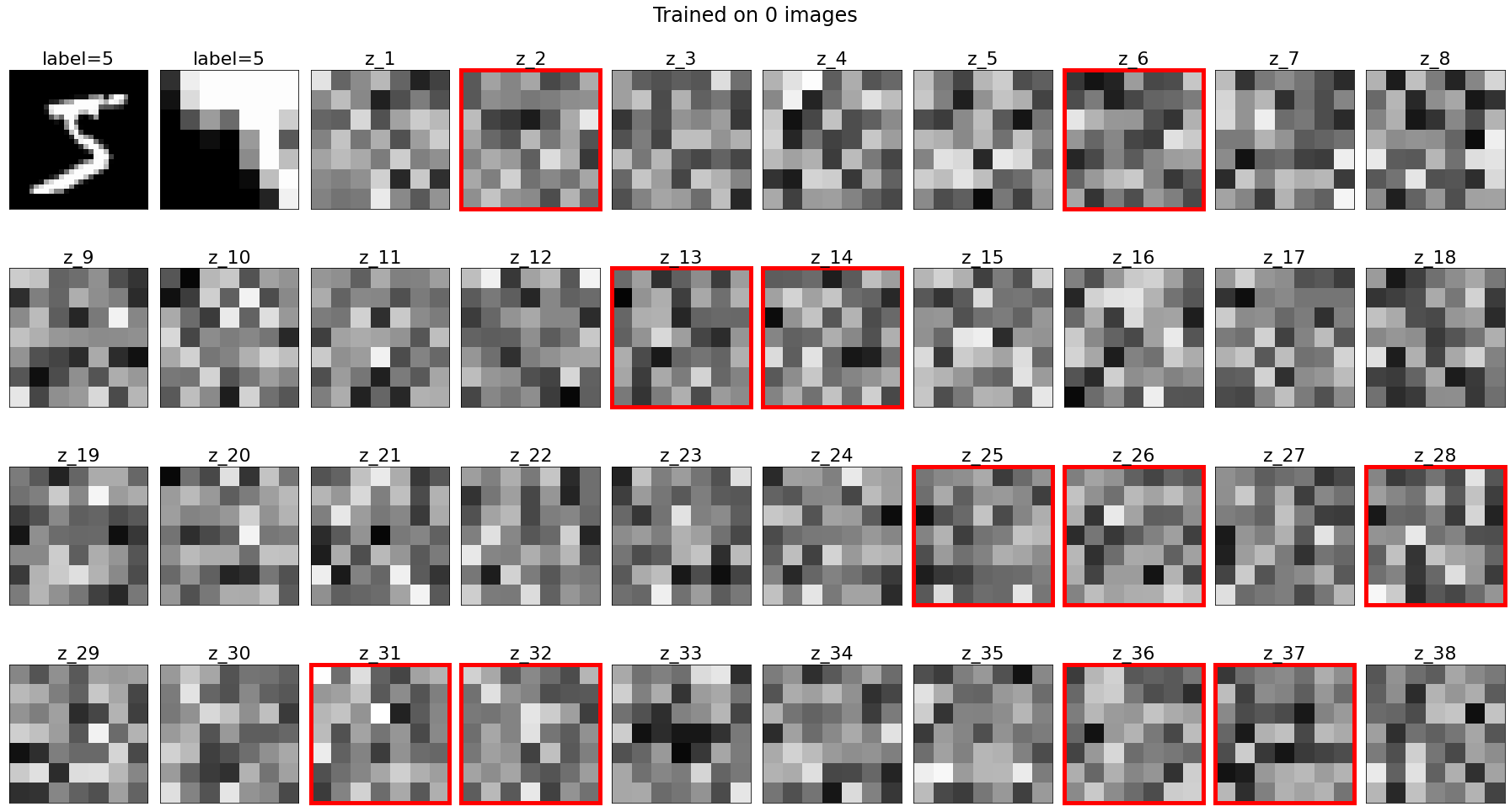}%
\caption{}%
\end{subfigure}\hfill%
\begin{subfigure}{.32\textwidth}
\includegraphics[width=\columnwidth]{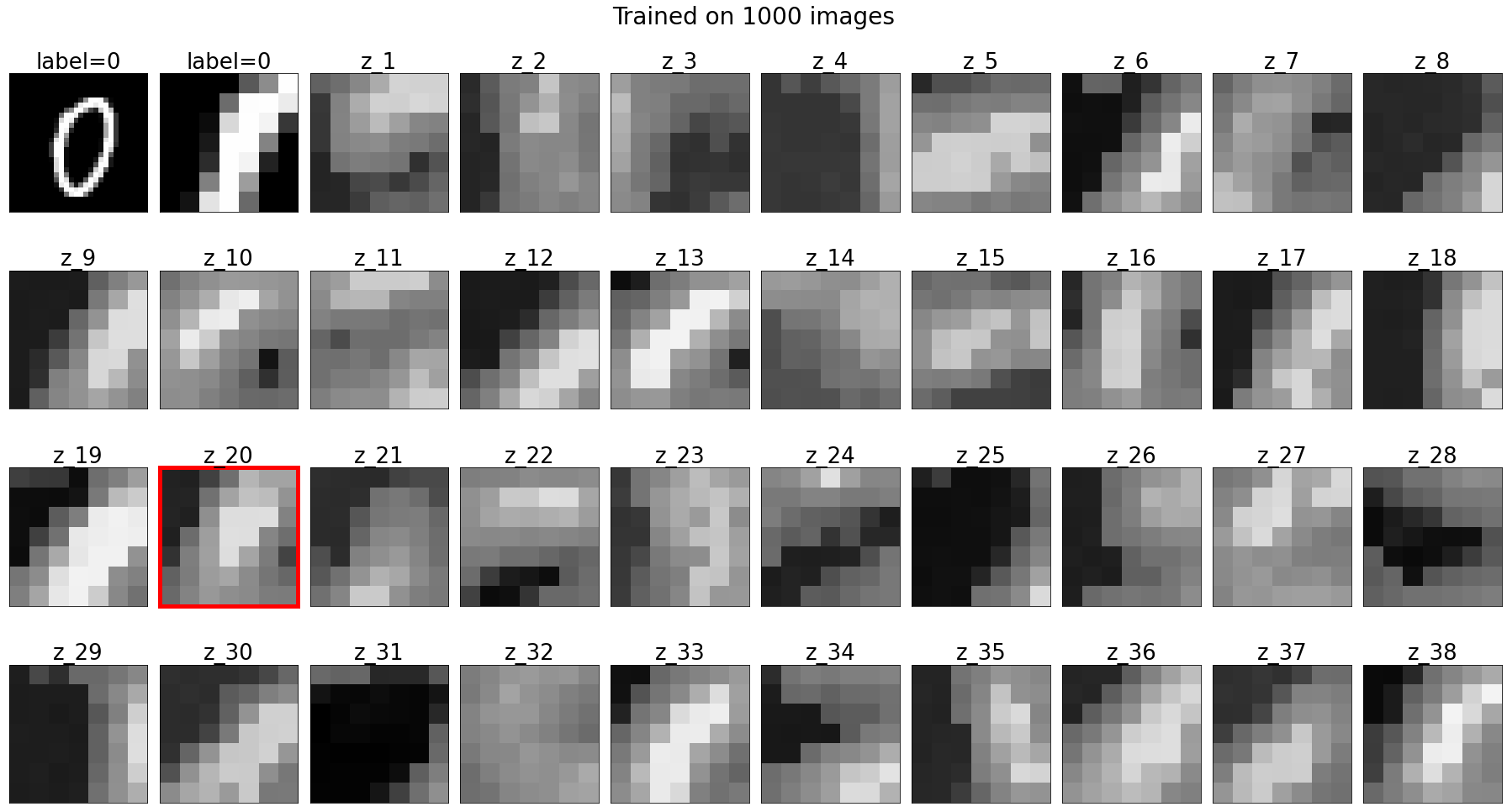}%
\caption{}%
\end{subfigure}\hfill%
\begin{subfigure}{.32\textwidth}
\includegraphics[width=\columnwidth]{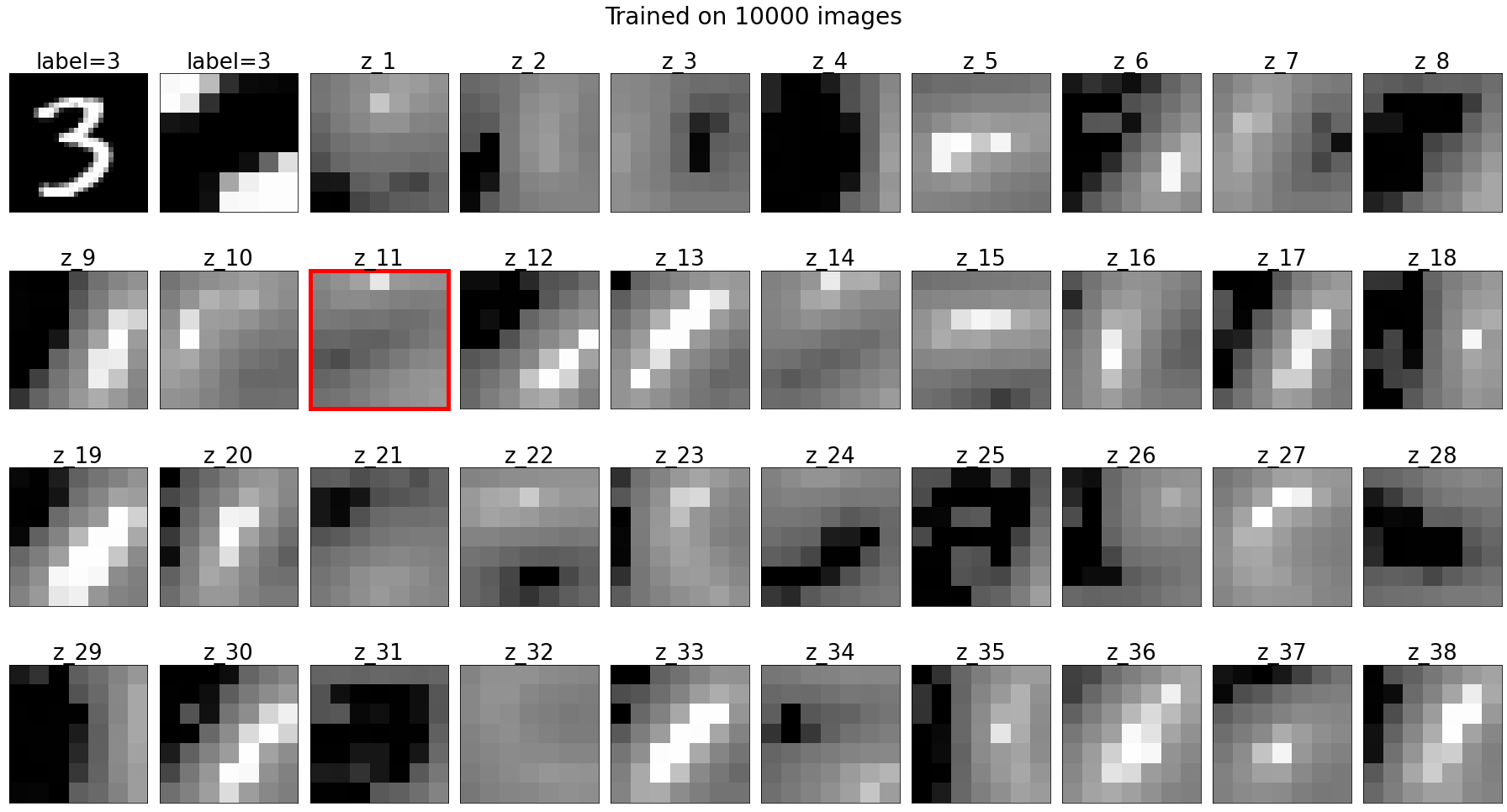}%
\caption{}%
\end{subfigure}\hfill%

\begin{subfigure}{.32\textwidth}
\includegraphics[width=\columnwidth]{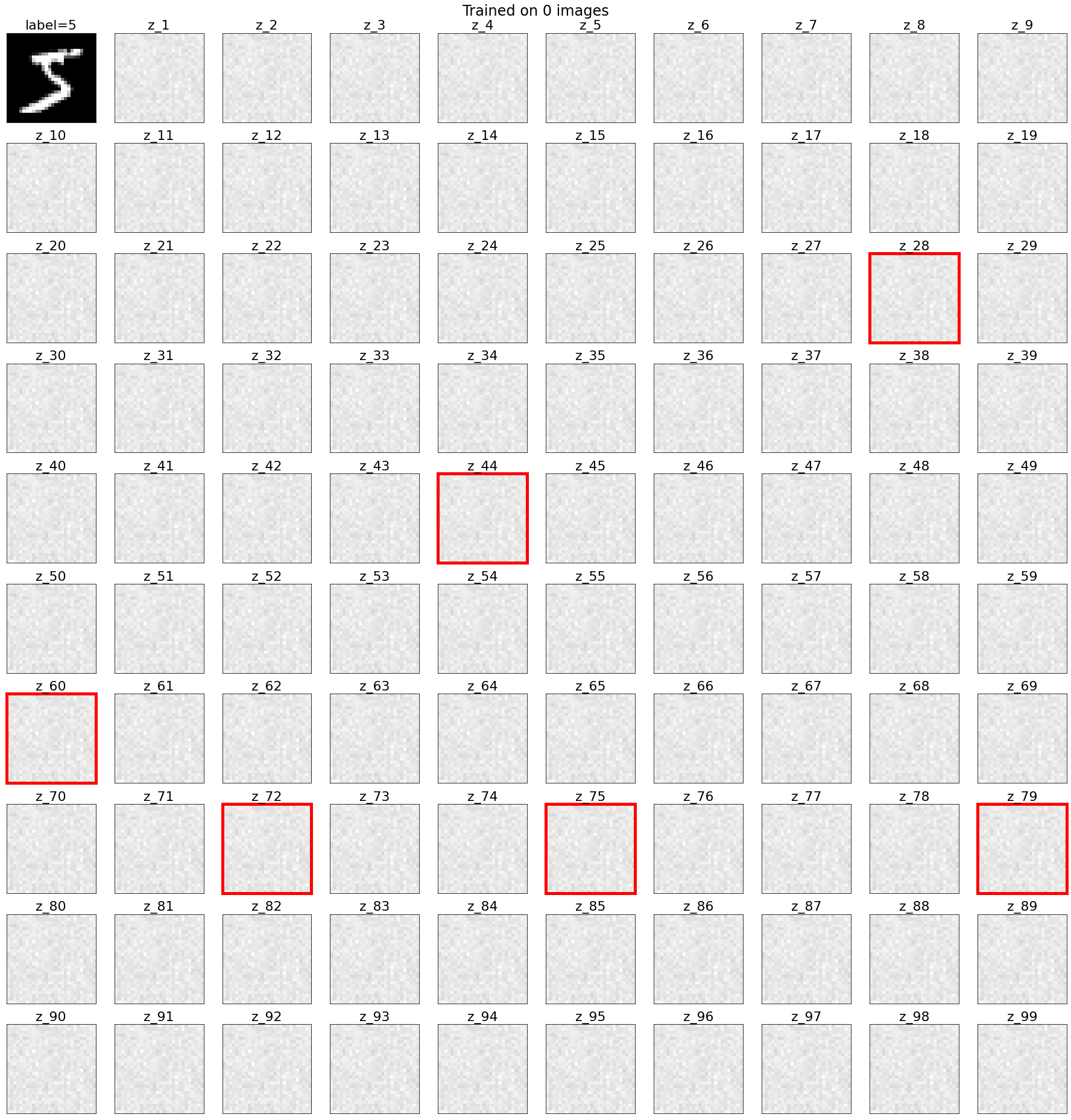}%
\caption{}%
\end{subfigure}\hfill%
\begin{subfigure}{.32\textwidth}
\includegraphics[width=\columnwidth]{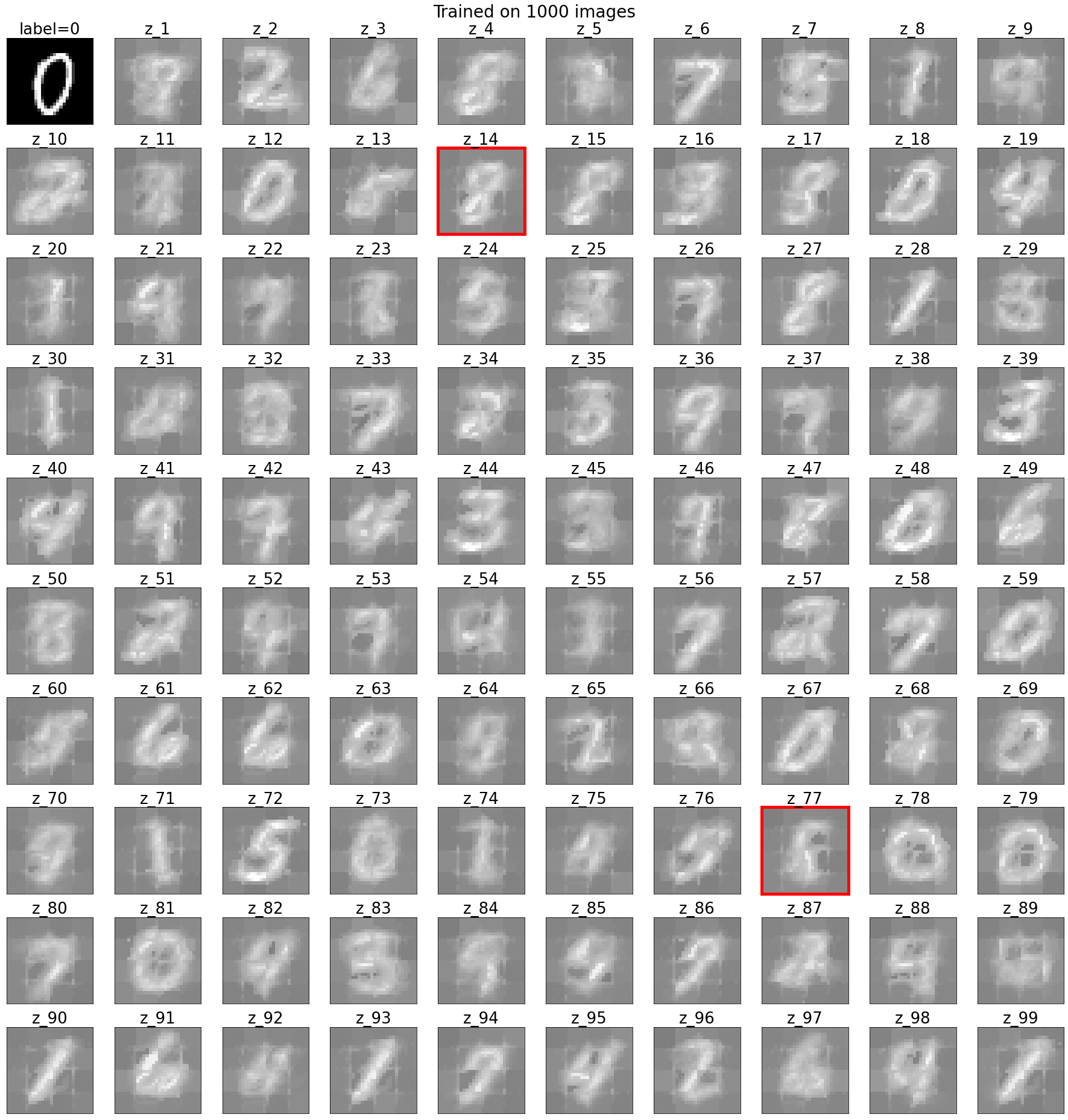}%
\caption{}%
\end{subfigure}\hfill%
\begin{subfigure}{.32\textwidth}
\includegraphics[width=\columnwidth]{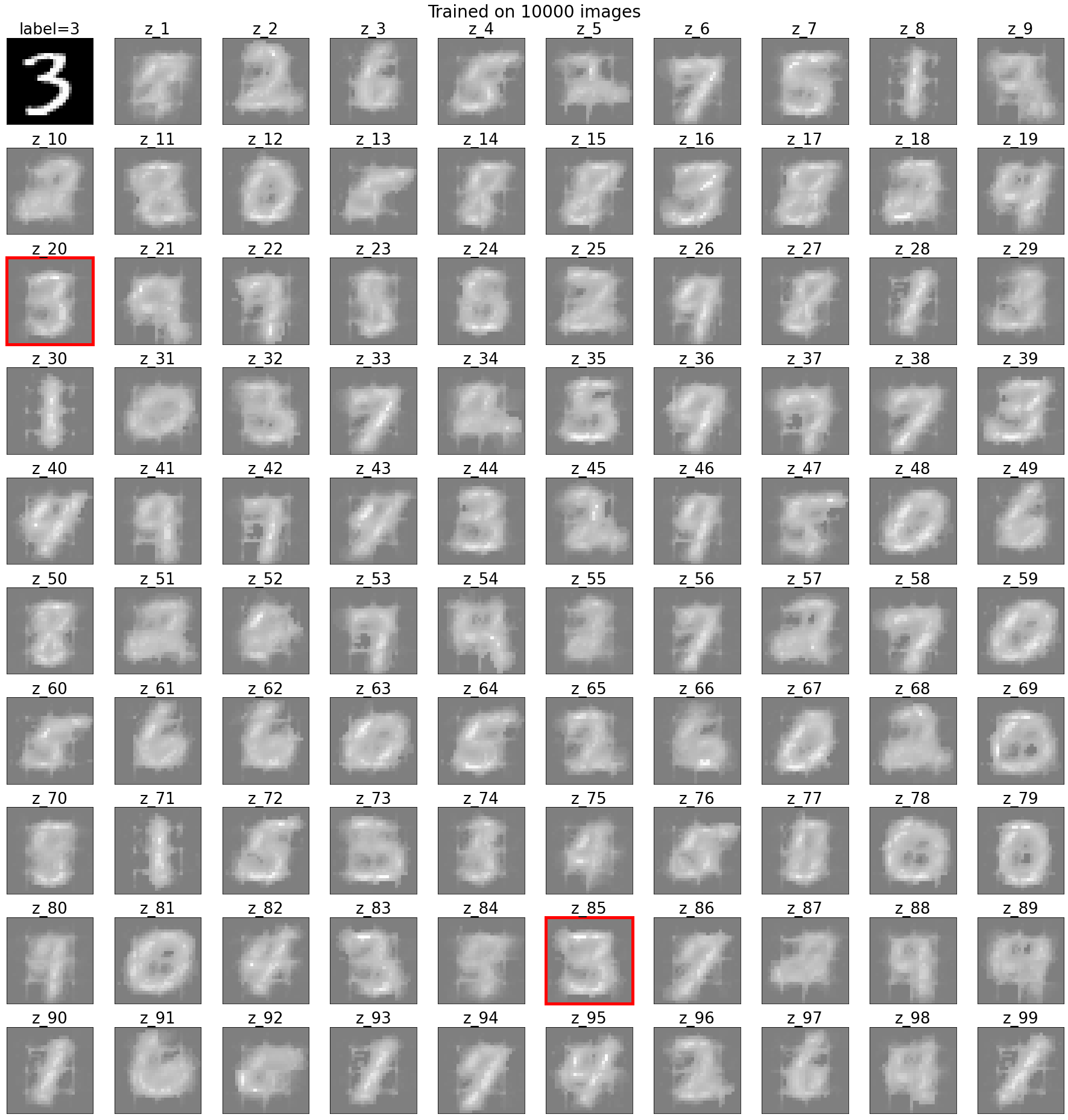}%
\caption{}%
\end{subfigure}\hfill%

\begin{subfigure}{.32\textwidth}
\includegraphics[width=\columnwidth]{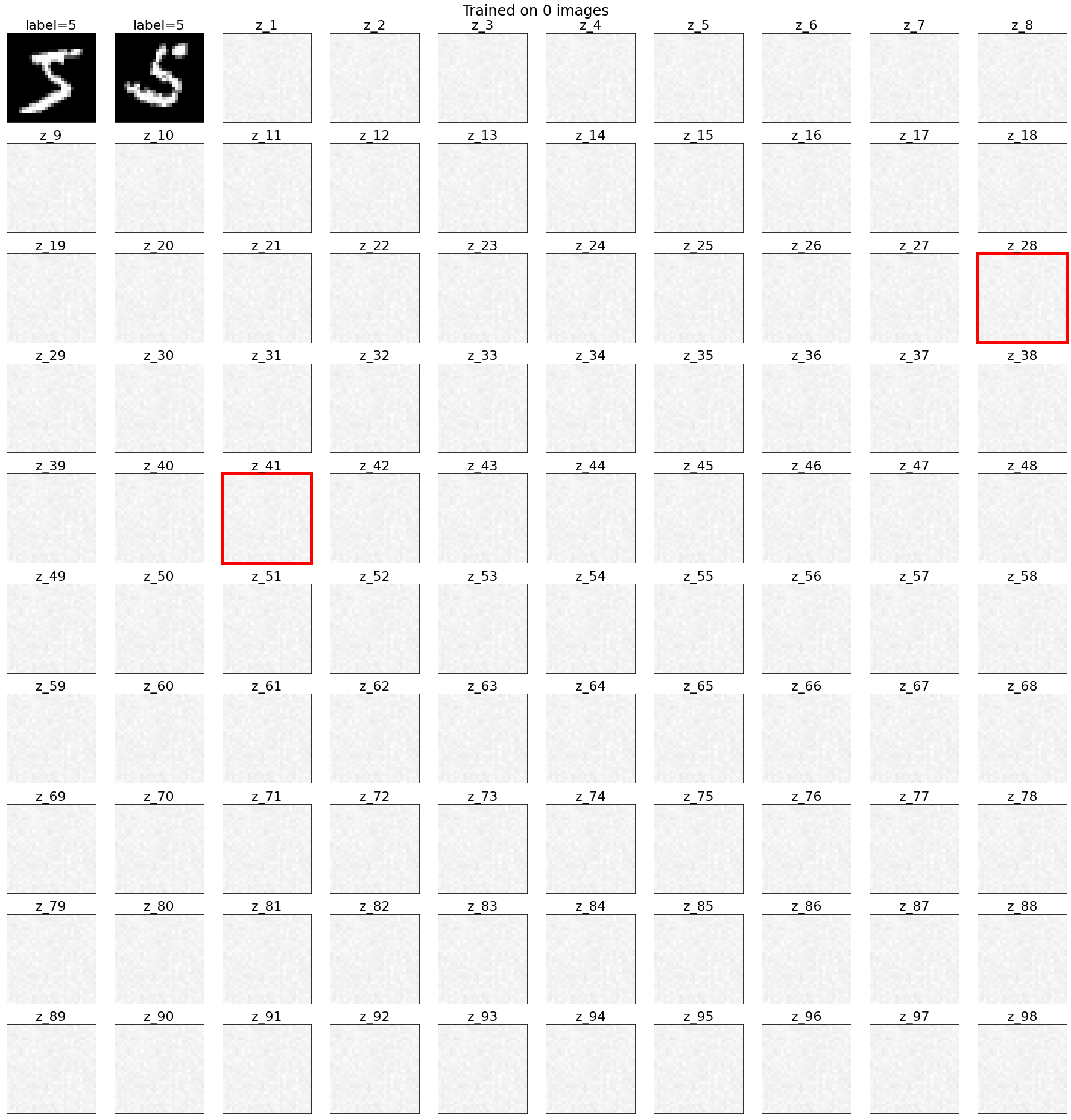}%
\caption{}%
\end{subfigure}\hfill%
\begin{subfigure}{.32\textwidth}
\includegraphics[width=\columnwidth]{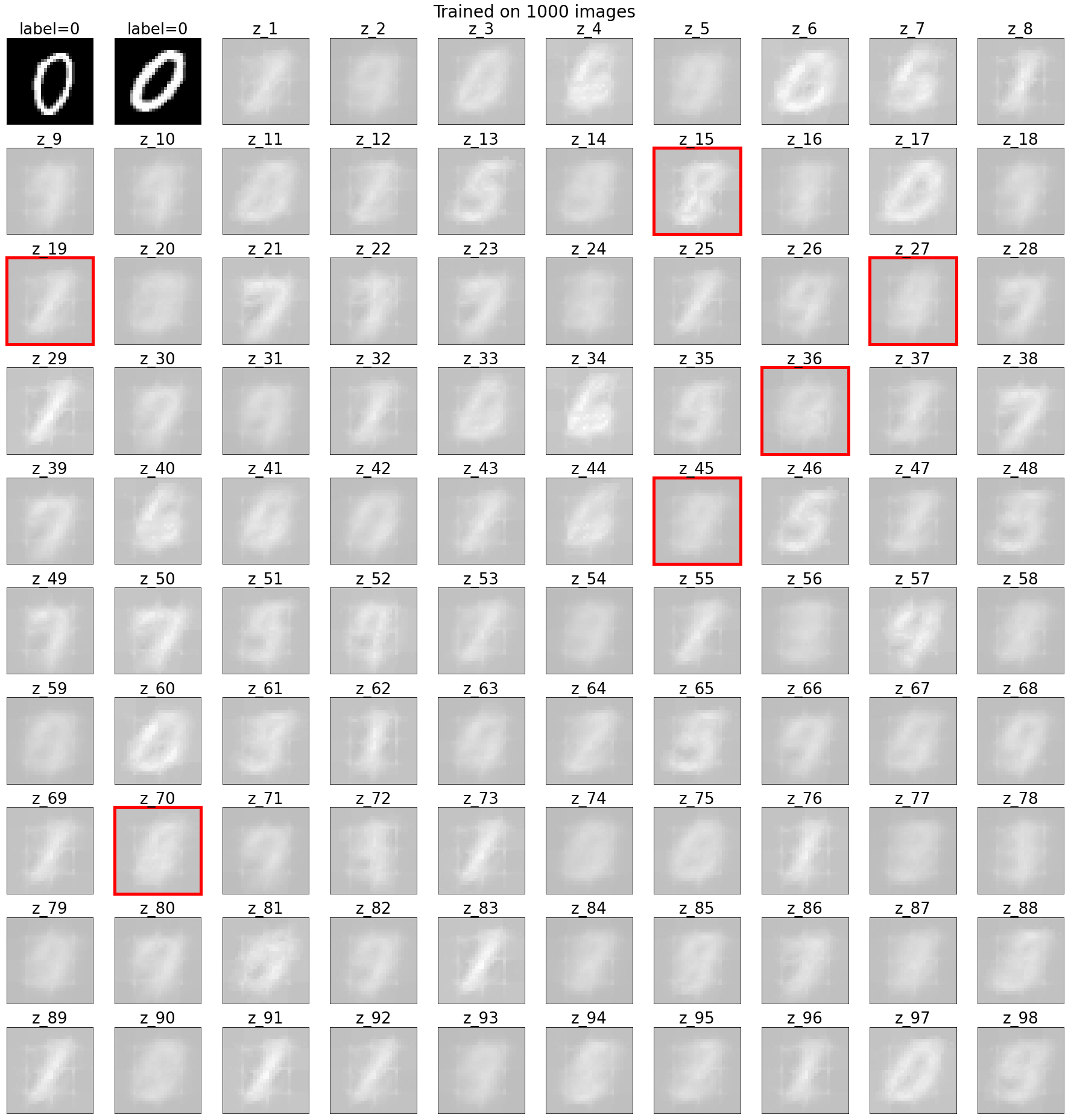}%
\caption{}%
\end{subfigure}\hfill%
\begin{subfigure}{.32\textwidth}
\includegraphics[width=\columnwidth]{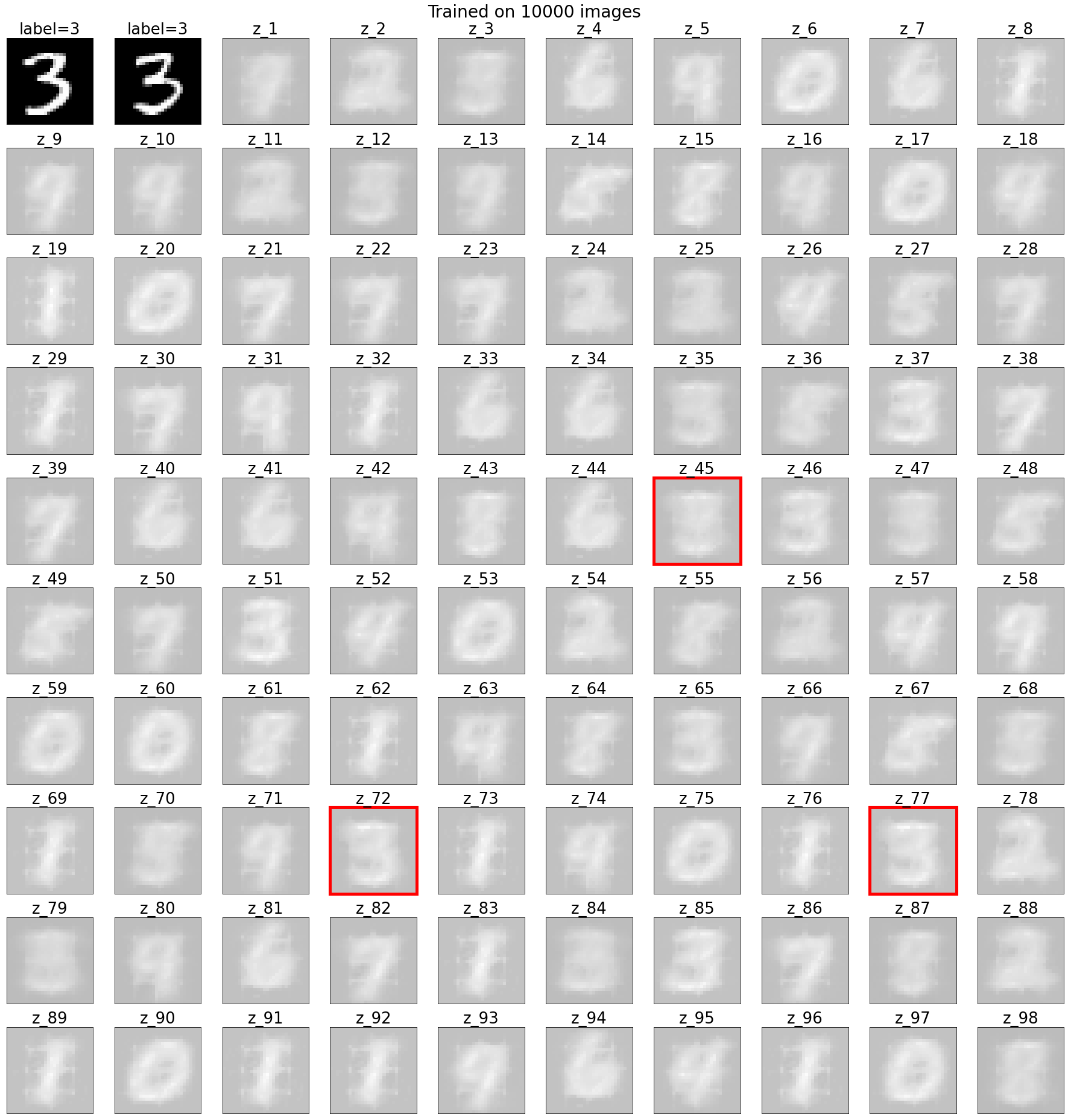}%
\caption{}%
\end{subfigure}\hfill%
\caption{Visualization of network weights after it has been exposed to respectively $0$, $1\,000$, and $10\,000$ stimuli (MNIST images). Within each sub-figure, the top left image(s) display an MNIST image that is being presented to the network, the remaining images are visualizations of the weights of all bottom-up connections that lead to a single neuron $z_k$. The red squares outline which neurons respond to the stimulus at hand.
(a), (b), and (c) visualize bottom-up connection weights going from a 7x7x2 cube of sensory neurons (that encode a 7x7 square of the image) to a single WTA circuit from the first layer of the WTA network. Of the two images in the top left of each sub-figure, the first image represents the digit being presented to the first layer as a whole, while the second image represents the part of the image being presented to the WTA circuit of which the connection weights are visualized. The figures show how WTA circuits in the first layer become sensitive to simple features such as particularly oriented lines and curves.
(d), (e), and (f) visualize bottom-up connection weights going from all sixteen WTA circuits in the first layer to a single WTA circuit of the second layer. The figures show how WTA circuits in the second layer form a relatively clear representation handwritten digits.
(g), (h), and (i) visualize bottom-up connection weights going from both WTA circuits in the second layer to a single WTA circuit of the third and final layer. The figures show how WTA circuits in the final layer form a slightly more generic representation of handwritten digits.}
\label{fig:weights-visualization}
\end{figure*}

When we assess the designs on the MNIST, this includes first of all a learning phase, where the network is exposed once to each of the $60\,000$ images that make up the MNIST training set. During this phase, the network weights evolve in an unsupervised manner according to STDP dynamics. After the learning phase, the network weights are frozen and performance is assessed against the full MNIST test set consisting of $10\,000$ images. Fig. \ref{fig:weights-visualization} visualizes how network weights evolve over time.

Performance of the network is assessed according to three measures: accuracy, confidence, and confidence error. Given that the network is trained in a completely unsupervised manner, it does not actually have an explicit notion of the underlying hidden variables. Thus, while the network may learn to distinguish between digits, it does not actually know that it is digits that it is distinguishing between. As such in order to determine classification accuracy we must determine what network output we correspond to which classification, which we do in the following fashion.

After training the network, we determine for each neuron in the final layer to which digit it responds the most (over the entire test set), this digit is then considered to be this neuron's classification. Then when a stimulus (MNIST image) is presented to the network, we observe which neurons respond, and can derive from this a distribution over digits (e.g. 8 spikes for digit zero, 2 spikes for digit one, ...). We consider the digit for which there were the most spikes to be the network's classification of the corresponding stimulus (where ties are broken at random). Further, we refer to the neurons generating the majority of spikes (in response to a specific stimulus) as the dominant neurons. Overall classification accuracy is the percentage of MNIST images that was classified correctly.

Secondly we include a measure of how confident the network is with respect to its classifications. As was just described, exposure to a stimulus may generate spikes that represent evidence for different digits. Any spike from a non-dominant neuron can be considered a sign of uncertainty, and thus the more spikes are distributed among non-dominant neurons, the less confident the network it is with regard to its classification. A network that performs well will both have a high confidence, and also be correct in its confidence. The former we define simply as the proportion of spikes generated by dominant neurons. The latter we can approximately measure by comparing the correctness of network classifications with the confidence the network has regarding said classifications. More concretely, over all classifications of a specific digit we would expect the proportion of spikes from non-dominant neurons to correspond to the proportion of wrong classifications. 

For example, consider all the times the network classifies an MNIST image from the test set as the digit zero. If 20\% of these images should actually have been classified as the digit nine, then we would expect that --- assuming the network is correct in its confidence --- on average classifications of the digit zero should consist for 20\% of spikes from neurons corresponding to the digit nine. We define the deviation from this expected value to be the confidence error.

Table \ref{tab:constants} summarizes the relevant constants, as well as the values to which we set these in our experiments. Further, the code used to perform the experiments is available on https://github.com/Grottoh/WTA-Network, and the corresponding data will soon be available on https://doi.org/10.34973/7fjz-va85.

\begin{table}[h]
\setlength{\tabcolsep}{3pt}

\caption{Experiment constants}
\begin{tabular*}{21pc}{p{28pt}p{44pt}p{164pt}}
\toprule
  
\multicolumn{3}{l}{}                                       \\

\underline{Constant} & \underline{Value} & \underline{Description} \\

$K_h$ & 38 & number of excitatory neurons per WTA circuit in layer one of the hierarchical and integration network \\
$K_o$ & 99 & number of excitatory neurons per WTA circuit in layer two of the hierarchical and integration network \\
$K_f$ & 98 & number of excitatory neurons in layer three of the integration network \\

--- & 150 & The number of discrete timesteps for which a stimulus (MNIST image) is presented to the network; one timestep is considered to be one millisecond \\
--- & 200 & The frequency (Hz) of the Poisson spike train generated by active sensory neurons \\

$\tau_f$ & 2 & STDP fast time constant \\
$\tau_s$ & 8 & STDP slow time constant \\

$\mu^{max}$ & $19.2558$ & maximum membrane potential of each neuron in \pmb{z} \\
$c$ & $1e{-}8$ & constant that plays a role in weight updates \\
--- & $-\ln c$ & maximum strength of connection weights of neurons $\pmb{z}$ \\

$\eta_k$ & $\dfrac{1}{N(z_k)^{0.8}}$ & adaptive learning rate weighing the weight update of neuron $z_k$; $N(z_k)$ represents the number of spikes generated by neuron $z_k$ over all stimuli at time $t$ \\

\bottomrule
\end{tabular*}
\label{tab:constants}
\end{table}

\subsection{Experiment 1: hierarchical vs integration}

\subsubsection{Results}

The first experiment compares the performance of the hierarchical design of \cite{guo_2019_hierarchical} against our integration design. We ran 10 experiments with the integration design. The integration design includes two hierarchical designs that function the same in the integration network as they would independently (in case of an absence of top-down processes). As a result, one assessment of the integration design (of $\mathcal{I}$) automatically includes two assessments of the hierarchical design (of $\mathcal{H}_a$ and of $\mathcal{H}_b$). We thus end up with two assessments of the hierarchical design, and one of the integration design, each averaged over 10 runs.

\begin{figure*}%
\centering
\begin{subfigure}{.33\textwidth}
\includegraphics[width=\columnwidth]{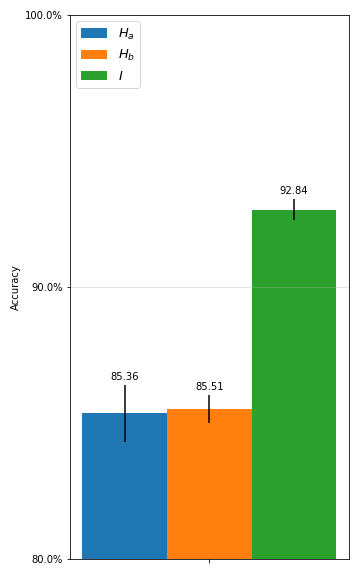}%
\caption{}%
\end{subfigure}\hfill%
\begin{subfigure}{.33\textwidth}
\includegraphics[width=\columnwidth]{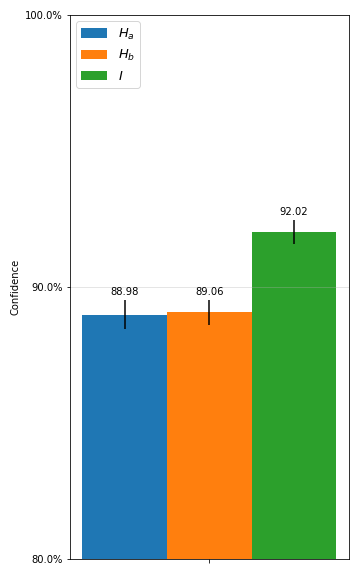}%
\caption{}%
\end{subfigure}\hfill%
\begin{subfigure}{.33\textwidth}
\includegraphics[width=\columnwidth]{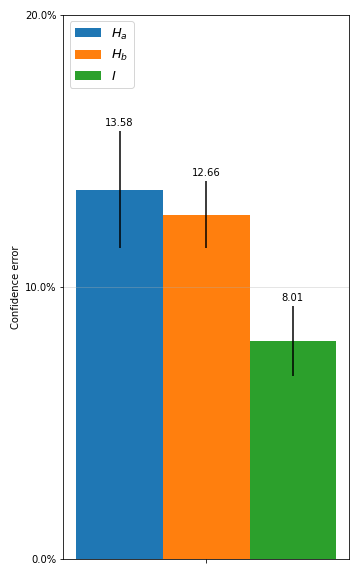}%
\caption{}%
\end{subfigure}\hfill%
\caption{Results for experiment 1. The figures display the overall (a) accuracy, (b) confidence, and (c) confidence error, averaged over 10 runs for two hierarchical networks ($\mathcal{H}_a$ and $\mathcal{H}_b$) and one integration network ($\mathcal{I}$).}
\label{fig:results-e1}
\end{figure*}

The hierarchical networks achieved an average accuracy of 85.36\% and 85.51\%, with standard deviations of 1.05\% and 0.52\% between runs. The integration network achieved an average accuracy of 92.84\% with a standard deviation of 0.39\% between runs. Further, the hierarchical networks achieved an average confidence of 88.98\% and 89.06\%, with standard deviations of 0.53\% and 0.47\% between runs. The integration network achieved an average confidence of 92.02\% with a standard deviation of 0.44\% between runs. Finally, the hierarchical networks achieved an average confidence error of 13.58\% and 12.66\%, with standard deviations of 2.15\% and 1.24\% between runs. The integration network achieved an average confidence error of 8.01\% with a standard deviation of 1.28\% between runs. These results are further displayed in Fig. \ref{fig:results-e1}.

\subsubsection{Interpretation}

Comparing the accuracy achieved by our replication of the hierarchical network of \cite{guo_2019_hierarchical} (85.36\% and 85.51\%) to the accuracy actually reported by \cite{guo_2019_hierarchical} themselves (84.89\%), we can see that these lie close to one another. This is as one would expect of a replication, and we expect that the slight difference that is present is likely a result of minor differences in implementation and hyperparameters.

Further, the results are in line with our hypothesis that WTA circuits can chain together separate WTA networks to improve inference and learning capacities. The integration network performs better than the hierarchical network on all our measures, displaying a higher accuracy, a greater confidence, and a lower confidence error. These results are expected, given that the integration network has access to more information.

The fact that WTA circuits are capable of processing probabilistic information represented by the spiking behaviour of other WTA circuits has several implications. Given that the manner in which information is encoded by a WTA circuit is independent of its source, these results suggest that WTA networks can be used to integrate information across multiple modalities (e.g. visual, haptic, auditory, ...). Further, hardware that can properly take advantage of the parallel and asynchronous nature of WTA networks (such as the human brain, or neuromorphic chips) can use these designs to process many stimuli at once and integrate information from these at low computational costs. 

To illustrate this, consider the costs associated with expanding the network. Lateral inhibition causes each WTA circuit to generate only a single spike at the time, and other inhibition mechanisms ensure that an appropriate firing rate of said neurons. This means that increasing the number of neurons per circuit will not increase the number of spikes generated by the circuit, and thus precludes the increased energy costs associated with an increase of spikes. Additionally, adding circuits to the same layer does not increase the processing time of this layer, given that all these circuits work in parallel with no direct dependence on one another. Indeed only the addition of layers increases the time it takes to propagate activation from sensory neurons to the final layer of the network, and that only by a single timestep. The main costs related to expanding a WTA network is thus related to the number of neurons and connections required. At least for the human brain this cost seems surmountable, given the many billions of neurons and the many trillions of connections that it consists of. As such, WTA networks appear to be a versatile tool for processing information in a Bayesian manner, and efficient with respect to speed and energy costs.

\subsection{Experiment 2: top-down processes}

\subsubsection{Results}

In the second experiment we investigate whether top-down processes can facilitate improved classification accuracy and confidence, and reduce confidence error. To this end we add sets of top-down connections between each layer, to mirror the bottom-up connections that are already in place. For the most part, the top-down connections function and evolve in the same manner as the bottom-up connections. The only difference present in our experiments, is their strength. Given that in our design the amount of firing neurons tends to decrease as we proceed to new layers in the network, bottom-up activation carries a greater combined weight than top-down activation. As such we include runs where we increase the strength of top-down processes by a scalar factor. Further, drawing upon literature that observe that top-down processes tend to increase in strength over time \cite{semedo_2021_feedforward}, we include runs where this is the case as well.

In order to properly illustrate the impact of top-down processes we report results of runs with several different parameters, which we label as follows. First of all runs with label $p\_no\mhyphen td$ concern runs where top-down processes are disabled (runs from experiment 1). Secondly, runs with labels $p\_td \stimes 1$, $p\_td  \stimes 2$, and $p\_td \stimes 3$ concern runs where top-down activation is strengthened by a constant factor of 1, 2, and 3 respectively. Finally, runs with label $p\_td \stimes \phi$ concern runs where top-down processes strengthen over time. During these runs, top-down signals from a neuron $z_k$ are strengthened by a factor $\phi_k(t) = \max(1.5+ 0.3 S(z_{k}, t)^{1.3}, 3)$ (a product of fine-tuning), where $S(z_{k}, t)$ is the number of spikes generated by neuron $z_k$ for the stimulus at hand at time $t$.

With respect to the impact of top-down processes we are interested in the following comparisons. First of all we want to know whether top-down processes are indeed capable of improving the performance of the integration network. And secondly, we are interested in the comparative performance of hierarchical networks trained as part of an integration network versus that of a hierarchical network trained in isolation.

\begin{figure*}%
\centering
\begin{subfigure}{.33\textwidth}
\includegraphics[width=\columnwidth]{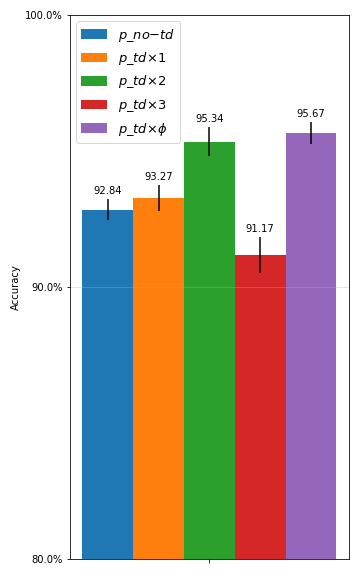}%
\caption{}%
\end{subfigure}\hfill%
\begin{subfigure}{.33\textwidth}
\includegraphics[width=\columnwidth]{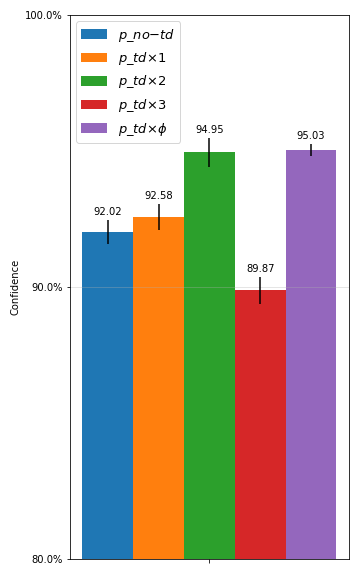}%
\caption{}%
\end{subfigure}\hfill%
\begin{subfigure}{.33\textwidth}
\includegraphics[width=\columnwidth]{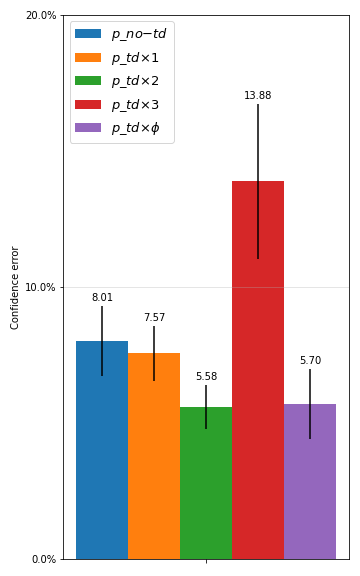}%
\caption{}%
\end{subfigure}\hfill%
\caption{Results for experiment 2 regarding the influence of top-down processes on integration network performance. The figures display the overall (a) accuracy, (b) confidence, and (c) confidence error, averaged over 10 runs for five integration networks. Experiments $p\_td \stimes 1$, $p\_td \stimes 2$, and $p\_td \stimes 3$ concern runs where the top-down activation is strengthened by a constant factor of 1, 2, and 3 respectively. Experiment $p\_td \stimes \phi$ concerns runs where top-down signal sent by a neuron $z_k$ is strengthened over time according to $\phi = \max(1.5+ 0.3 S(z_{k}, t)^{1.3}, 3)$, where $S(z_{k}, t)$ is the number of spikes generated by neuron $z_k$ for the stimulus at hand at time $t$. Results for the integration network of experiment 1 ($p\_no\mhyphen td$; without top-down processes) are included for reference.}
\label{fig:results-e2}
\end{figure*}

The results for the first comparison are displayed in Fig. \ref{fig:results-e2}. These results concern integration network accuracy, confidence, and confidence error under the aforementioned parameters, averaged over ten runs. The results show that top-down processes can indeed improve performance. In particular, runs $p\_td \stimes 2$ and $p\_td \stimes \phi$ (which perform roughly the same) show superior performance on all measures.

\begin{figure*}%
\centering
\begin{subfigure}{0.75\textwidth}
\includegraphics[width=\columnwidth]{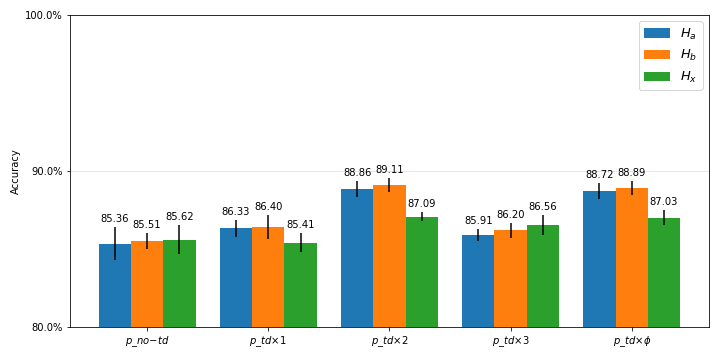}%
\caption{}%

\end{subfigure}\hfill%
\begin{subfigure}{0.75\textwidth}
\includegraphics[width=\columnwidth]{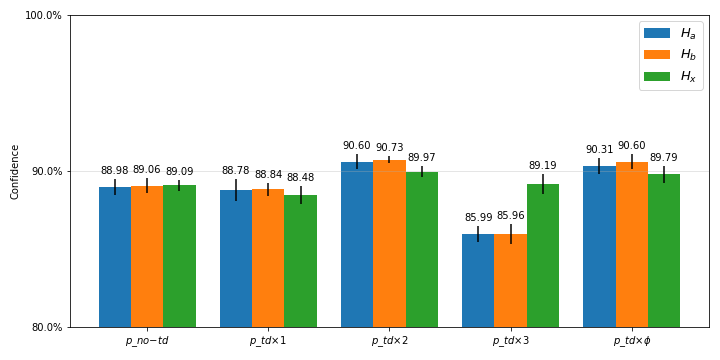}%
\caption{}%

\end{subfigure}\hfill%
\begin{subfigure}{0.75\textwidth}
\includegraphics[width=\columnwidth]{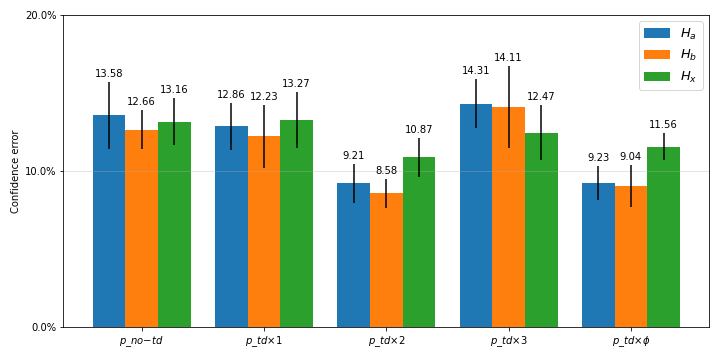}%
\caption{}%
\end{subfigure}\hfill%
\caption{Results for experiment 2 regarding the influence of top-down processes on hierarchical network performance. Comparison of hierarchical networks trained with top-down processes enabled. The figures display the overall (a) accuracy, (b) confidence, and (c) confidence error, averaged over 10 runs for three hierarchical networks. Hierarchical networks $\mathcal{H}_a$ and $\mathcal{H}_b$ were trained as part of an integration network, thus each receiving feedback from the final layer of said integration network. In contrast, hierarchical network $\mathcal{H}_x$ was trained in isolation.}
\label{fig:results-e2-h-comparison}
\end{figure*}
network
The results for the second comparison are displayed in Fig. \ref{fig:results-e2-h-comparison}. These results show that the hierarchical networks that are trained as part of an integration network ($\mathcal{H}_a$ and $\mathcal{H}_b$) gain a greater boost in performance as a result of top-down processes than the independently trained hierarchical network ($\mathcal{H}_x$). As is the case for the integration network, the gain in performance is the greatest for all networks under parameters $p\_td \stimes 2$ and $p\_td \stimes \phi$ (for which the results are roughly the same). Under these parameters the performance gain is greatest for $\mathcal{H}_a$ and $\mathcal{H}_b$ on all measures.

\subsubsection{Interpretation}

The results show that top-down processes can be used to improve network performance on all of our measures. Simply including top-down connections that mirror the bottom-up connections, but otherwise function and evolve in the same manner ($p\_td \stimes 1$), is enough to improve classification accuracy and confidence, and reduce confidence error. Further, multiplying the strength of top-down signals can further improve performance on these measures ($p\_td \stimes 2$), though this does have its limits ($p\_td \stimes 3$). Increasing the strength of top-down processes over time also yields peak performance ($p\_td \stimes \phi$), though the improvement with respect to simple signal strength multiplication ($p\_td \stimes 2$) seems negligible.

The improvement is a result of the more confident circuits in the network informing the less confident ones. More confident circuits will have a more stable spike output, meaning that a smaller variety of neurons respond to the stimulus at hand. As such, the activation that it propagates forward and backward is more focused on the strong connections of these specific neurons. On the other hand, a completely uncertain circuit where each neuron fires as often as the other, will propagate a uniform signal forward and backward. Therefore, the more confident circuit, with the more focused signal, will come to exert a more distinguished influence on the rest of the network, a process that strengthens itself over time (thus spreading confidence throughout the network). This accounts for the improved performance that we find.

The results finally showed that top-down processes improve learning throughout the WTA network. Hierarchical networks that were trained as a part of an integration network (with top-down processes enabled), learned to perform better in isolation (i.e. while not being part of an integration network) than an otherwise identical hierarchical network that was trained in isolation. These results suggest that top-down processes facilitate learning by transferring information between sub-networks that would otherwise be completely independent from one another; this is in line with literature that points out that top-down processes play a role in encoding and recall of information \cite{gilbert_2013_top-down}. Further, this implies that the role of top-down processes become more prominent as the network grows larger.

\section{Conclusion}\label{sec:conclusion}

We have shown that a WTA circuit is capable of integrating the probabilistic information represented in multiple WTA networks, and that this improves the network's capacity to infer hidden variables. Moreover, we have shown that in multi-layered WTA networks, top-down processes further improve the performance of said networks. These findings fit with what we know of the brain. Our findings confirm the notion that WTA networks are suitable processing components for the integration of information, being capable of representing generative probabilistic models, inferring the hidden variables in such a model, and approximately learning the parameters of such a model. Further, it is able to do this with a representation unspecific to any modality, suggesting that it is capable of integrating information from different senses, and can be seen as an abstract concept representation. 

Our finding that top-down processes have a beneficial influence on network performance is in line with observations of how the brain behaves \cite{gilbert_2013_top-down, harris_2013_cortical, dijkstra_2017_distinct, semedo_2021_feedforward}. Though earlier schools of thought expected activation in the brain to be a bottom-up process, recently there has been a shift to the belief that processing in the brain is a combination of bottom-up and top-down processes. We have shown that this latter school of thought fits with the dynamics of WTA networks, and have given an explanation for the observed effect. 
It is noted in \cite{semedo_2021_feedforward} that the role and impact of top-down processes become greater as the task becomes more complex. This too fits with our finding that top-down processes facilitate the sharing of learned information between parts of the network that cannot interact with one another through bottom-up processes alone; implying that top-down processes help to spread learned information throughout the network, where a larger network that processes more sources of information has the greater potential to be spread learned information.

The algorithm proposed here can --- given the proper neuromorphic hardware --- operate completely according to a number of the most important neuromorphic principles, being capable of functioning with local, event-based, parallel, and asynchronous computations. The point of this research is not to impress with the inference capabilities of this method, but rather to show that it is able to perform approximate Bayesian inference whilst adhering to these principles. Further, we want to emphasize the experimental finding that we can expand this network by the simple addition of layers, thereby allowing the processing of additional stimuli, and can via feedback allow otherwise separate layers to communicate with and learn from one another. Given the parallel and asynchronous nature of the algorithm this can --- again given the proper hardware (such as perhaps the human brain) --- be done with little to no additional processing time. The event-based and sparse nature of the algorithm further potentially allow for extremely energy-efficient execution. The greatest cost related to expanding networks is the requirement of the extra neurons and connections. This may be a challenge for neuromorphic hardware in the near future, but at least in the human brain we have an example of 'neuromorphic hardware' with plenty of neurons and connections, showing that this is not an insurmountable obstacle.

Future research in this area has several challenges to tackle. For one it will be necessary to move to more complex stimuli than MNIST digits. Further, considering more closely how to encode the stimuli will be vital. The encoding used in the present research violates several key neuromorphic principles in that it is neither sparse, nor event-based. Future research might instead encode visual data in a manner similar to for example a DVS. Additionally, in the future it will be interesting to extend this method to work with temporal data such as speech, and to see it implemented on neuromorphic hardware.

Future research should be careful when comparing this method to traditional machine learning approaches. It will be tempting to compare it directly with e.g. a Convolutional Neural Network (CNN) when performing visual classification tasks, yet such a comparison may not fit. The danger is that with such comparisons the focus becomes fine-tuning and tweaking the approach to improve its performance on a small set of common baseline tasks (e.g. first MNIST, and then more complex image classification tasks). While this could indeed improve the method, we argue that the true potential of this approach lies not with optimized performance on very narrowly defined tasks such as traditional image classification, but with more complex and general tasks such as are more common in a real-world setting. For example, when a human attempts to 'classify' a digit they often have an abundance of additional information. They may for instance know that the number is even, because it is a house number on a particular side of the street. If they are unsure about their 'classification', they may additionally act to change the environment (e.g. by wiping off some dirt) or to change their perspective of it (by looking at it from a closer distance). This integration of information, and these actions taken, require processing in different areas of the overall network, and communication between said areas. In order to do this in an quick and energy-efficient manner the event-based, parallel, and asynchronous nature of the algorithm proposed here will be useful, or even essential. Therefore we advice that future research keeps this in mind when designing tasks, and making performance comparisons.

\bibliographystyle{IEEEtran}
\bibliography{references.bib}

\begin{IEEEbiography}{Otto van der Himst}{\space} received the B.Sc. degree in Psychology from the Radboud University Nijmegen in 2014. Further, he received the B.Sc. degree in Artificial Intelligence from the Radboud University Nijmegen, in 2017. Here he is currently pursuing the M.Sc. degree Artificial Intelligence, and the M.Sc degree in Data Science. His current research interests include deep learning, reinforcement learning, and the development of SNN algorithms for neuromorphic architectures.
\end{IEEEbiography}

\begin{IEEEbiography}{Leila Bagheriye}{\space}(M20) received the B.Sc. degree in electrical engineering from the University of  Tabriz, Tabriz, Iran, in 2010 and the  M.A.Sc. degree in electronics engineering with honor from the University of Zanjan, Zanjan, Iran, in 2012, She received her  Ph.D. degree in electronics engineering in 2018 and she was a visiting Ph.D. student at the ICE- Laboratory, Aarhus University, Aarhus, Denmark. From 2019-2021 she was with the CAES group at the University of Twente and focused on machine learning based reliability enhancement of multi-sensori platforms. Since 2021 she is with the Donders Institute at Radboud University where she focuses on implementation of neuromorphic algorithms emphasizing on Bayesian networks and spiking neural networks.
\end{IEEEbiography}

\begin{IEEEbiography}{Johan Kwisthout}{\space}is associate professor in Artificial Intelligence and principle investigator at the Donders Institute at Radboud University, where he leads the Foundations of natural and stochastic computing group. He holds M.Sc. degrees in computer science and artificial intelligence and received his Ph.D. degree in computer science in 2009 from Utrecht University. He is interested in the foundations and applications of probabilistic graphical models, particularly computations on Bayesian networks, as well as their realization in neuromorphic architectures.
\end{IEEEbiography}

\end{document}